\documentclass[letterpaper]{article} 
\usepackage{aaai24}  
\usepackage{times}  
\usepackage{helvet}  
\usepackage{courier}  
\usepackage[hyphens]{url}  
\usepackage{graphicx} 
\usepackage{natbib}  
\usepackage{caption} 
\usepackage{algorithm}
\usepackage{algorithmic}
\usepackage{multirow}
\usepackage{tabularx}

\usepackage{fancyhdr,graphicx,amsmath,amsthm, amssymb, mathtools, scrextend, enumitem}
\newtheorem{definition}{Definition}
\newtheorem{theorem}{Theorem}
\newtheorem{lemma}[theorem]{Lemma}

\usepackage{booktabs,makecell}

\usepackage{newfloat}
\usepackage{listings}
\DeclareCaptionStyle{ruled}{labelfont=normalfont,labelsep=colon,strut=off} 
\floatstyle{ruled}
\newfloat{listing}{tb}{lst}{}
\floatname{listing}{Listing}
\title{Understanding Distributed Representations of Concepts \\ in Deep Neural Networks without Supervision}
\author{
    Wonjoon Chang\textsuperscript{\rm 1} \equalcontrib,
    Dahee Kwon\textsuperscript{\rm 1} \equalcontrib,
    Jaesik Choi\textsuperscript{\rm 1, \rm 2} 
}

\affiliations{
    
    \textsuperscript{\rm 1} Korea Advanced Institute of Science and Technology \\
    \textsuperscript{\rm 2} INEEJI\\  
    $\{$one\_jj, daheekwon, jaesik.choi$\}$@kaist.ac.kr

}

\begin{document}

\maketitle

\begin{abstract}
Understanding intermediate representations of the concepts learned by deep learning classifiers is indispensable for interpreting general model behaviors. Existing approaches to reveal learned concepts often rely on human supervision, such as pre-defined concept sets or segmentation processes. In this paper, we propose a novel unsupervised method for discovering distributed representations of concepts by selecting a principal subset of neurons. Our empirical findings demonstrate that instances with similar neuron activation states tend to share coherent concepts. Based on the observations, the proposed method selects principal neurons that construct an interpretable region, namely a Relaxed Decision Region (RDR), encompassing instances with coherent concepts in the feature space. It can be utilized to identify unlabeled subclasses within data and to detect the causes of misclassifications. Furthermore, the applicability of our method across various layers discloses distinct distributed representations over the layers, which provides deeper insights into the internal mechanisms of the deep learning model.
\end{abstract}

\section{Introduction}

Despite the remarkable performance of Deep Neural Networks (DNNs) in learning intricate data relationships~\citep{lecun2015deep}, their inherent lack of transparency remains a significant challenge. This opacity makes it difficult to understand the decision-making process, reducing model reliability and weakening the applicability in risk-sensitive domains where careful decisions are needed~\cite {gunning2019xai,samek2019explainable}. To gain insights into the general behaviors of DNNs, it is essential to reveal the semantic representations that DNNs learn. Our primary goal is to understand the distributed representations of concepts embedded within a trained model without external supervision. This approach facilitates the identification of diverse concepts within the model, including subclass distinctions, class-agnostic concepts, and even concepts that might contribute to misclassification.

\begin{figure}[t]
\centering
\includegraphics[width=0.95\columnwidth]{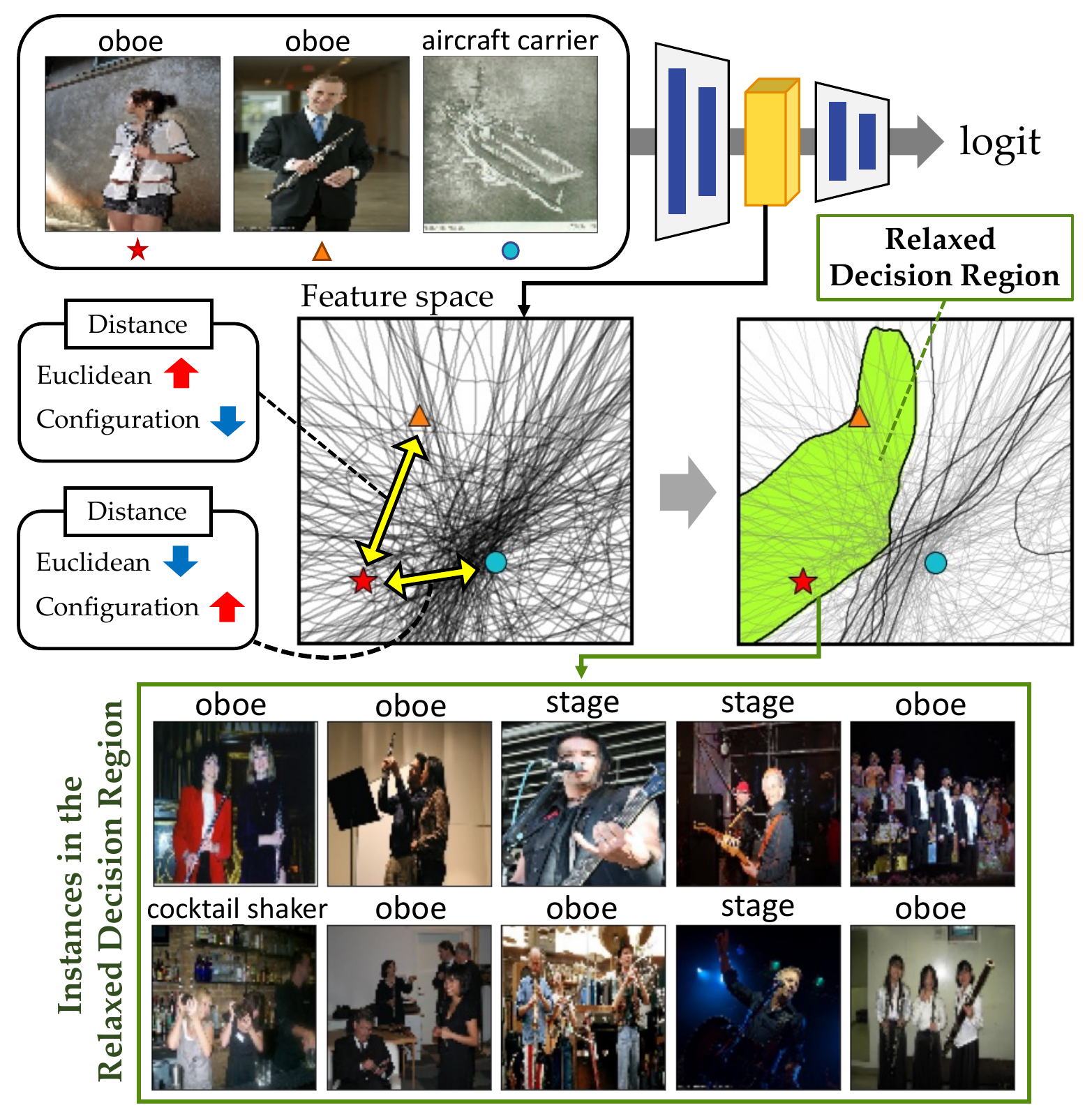}
\caption{\normalsize Relaxed Decision Region (RDR). Top: Description of the target sample and its neighbors. Middle: Visualization of the feature space and the RDR. Our RDR framework groups instances that have similar neuron activation states in the feature space. Bottom: Instances in the RDR share the coherent concept of `a person with a stick'.}
\label{fig:main-concept}
\end{figure}

\begin{figure*}[ht!]
\centering
\centerline{\includegraphics[width=0.85\textwidth]{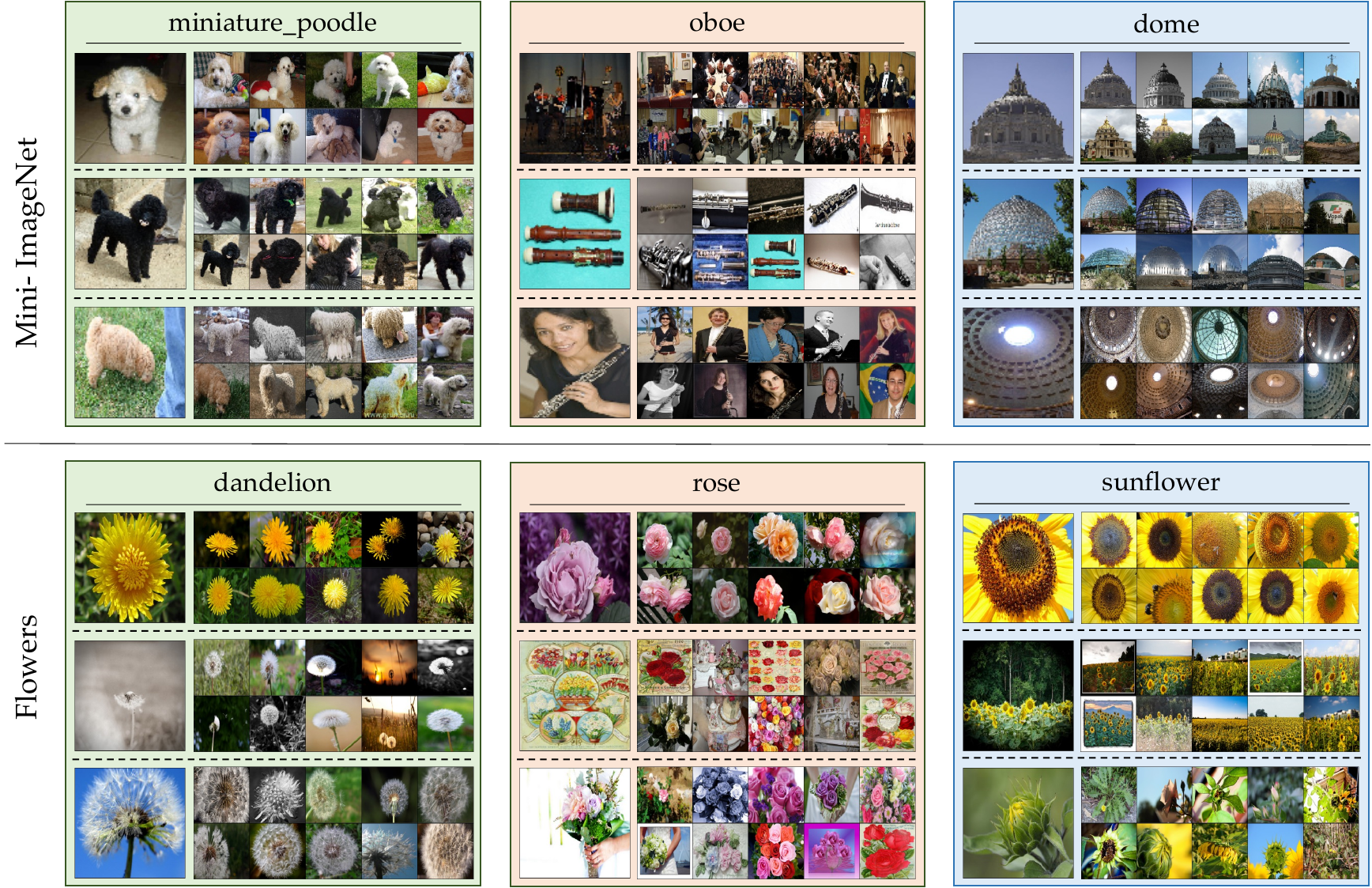}}
\caption{\normalsize Without prior knowledge of label information, our RDR framework successfully captures learned concepts such as subclasses, shapes, crowds,
composition, and the degree of flowering, as well as simple color schemes.}
\label{fig:main-result}
\end{figure*}

Various eXplainable Artificial Intelligence (XAI) methods have been developed to enhance the transparency of a model. The gradient-based methods reveal which parts of input significantly contribute to the model's classification result based on the gradient information~\citep{simonyan2014visualising, bach2015pixel, selvaraju2017grad, sundararajan2017axiomatic, chattopadhay2018grad}. Yet, they focus on individual instances rather than the representative concepts that the model has learned in terms of generality. In this context, concept-based explanation methods emerge to provide more nuanced and general explanations~\citep{kim2018interpretability, ghorbani2019towards, crabbé2022concept}. Despite their perceptually intuitive results, most methods heavily rely on human supervision. Not only does it take substantial cost to require a refined concept dataset, but also there is no guarantee that the pre-defined concepts truly reflect the model behavior.

Another line of research takes a distinct approach that directly discloses the concepts embedded in the model by observing the role of internal neurons~\cite{bau2017network,fong2018net2vec}. In particular, under the assumption of a \textit{distributed representation}~\cite {hinton1984distributed}, \citet{fong2018net2vec} find combinations of neurons that represent learned concepts with segmentation information. Nevertheless, they still require human-annotated information and often entail computationally intensive manual searching.

In this paper, we present an interpretation framework aiming to elucidate learned concepts in DNNs. The proposed framework captures concept representations by leveraging the inherent information in the intermediate layers and offers example-based explanations without supervision. We first empirically demonstrate that instances with similar activation states share coherent concepts, and select a subset of relevant neurons, namely the \textit{principal configuration}. Using the \textit{principal configuration}, our approach constructs an interpretable region named \textit{Relaxed Decision Region} (RDR) (Figure~\ref{fig:main-concept}). Our RDR framework can reveal various learned concepts, including subclasses (Figure~\ref{fig:main-result}), concepts leading to misclassification, and diverse concepts across different layers.

\section{Related Work}
\label{sec:relatedwork}

Concepts learned by DNNs can be unveiled by leveraging human supervision directly. Concept discovery methods, when given a predefined concept set, compute a relevant concept vector or a region in the internal feature space~\citep{kim2018interpretability,schrouff2021best,crabbé2022concept,sajjad2022neuron,oikarinen2022clip}. Although attempts have been made to bypass the need for pre-defined concept sets~\citep{ghorbani2019towards,kusters2020conceptual,koh2020concept}, they entail other types of costs, such as the segmentation process.

Other notable approaches aim to identify the role of internal components in DNNs, such as convolutional filters, by aligning activation patterns with pre-defined information~\citep{bau2017network,fong2018net2vec,angelov2020top2vec,achtibat2022towards}. These approaches commonly utilize segmentation information to evaluate individual concepts captured through internal neurons. Recently, \citet{oikarinen2022clip} introduced a method providing textualized explanations for internal neurons, leveraging a CLIP model with pre-defined textual concept sets. Although these approaches still involve human supervision, they offer strong empirical evidence that internal neuron activations potentially encode information about learned semantics.

To avoid the requirement of supervision, example-based explanation methods select representative exemplars that summarize data distribution~\citep{kim2016examples, khanna2019interpreting, cho2021interpreting}. Despite the advantages of their unsupervised algorithms, there is no guarantee that the exemplars precisely reflect the decision logic of the model. Another way is to explicitly design a specific structure capable of learning prototypical representations itself~\citep{chen2019looks, nauta2023pip}. In such cases, however, the model structure is necessarily constrained.

\citet{lam2021finding} proposed the Representative Interpretation (RI) method, which establishes a region in the feature space encompassing the maximum number of instances of a target class. While RI focuses on a specific target class, our objective is to construct a region where instances share coherent concepts without being constrained by class distinctions. We achieve it by leveraging activation patterns in an unsupervised manner. The idea is based on the fact that activation patterns are closely linked to model decisions~\citep{chu2018exact, gopinath2019property} and captured representations~\citep{bau2017network,fong2018net2vec}.

\section{Problem Definition}
\label{sec:problem}
 
To set the groundwork for our discussion, this section introduces necessary terminologies and definitions. We start by clarifying the key properties of the concept sets we aim to find from a trained DNN. (1) \textbf{\textit{Learned representation}}: The concept should originate within the model and be extractable from the set of neurons.  (2) \textbf{\textit{Coherence}}: The concept should convey a common semantic meaning that is consistently observed across multiple instances. (3) \textbf{\textit{Discrimination}}: The extracted concept set should be distinguishable from non-concept sets, ensuring straightforward human comprehension. 

To identify a concept with the aforementioned properties, we select principal neurons that represent the corresponding concept, enabling interpretation for a target instance. The chosen neurons constitute the \textit{Relaxed Decision Region}, described in the next section. The motivation of our neuron-selection approach is rooted in the principle of distributed representation~\cite{ hinton1986learning, fong2018net2vec}, suggesting that a model's learned concept representations are distributed across multiple internal components, akin to neurons in DNNs.

\subsection{Terminologies}

Consider a neural network $F:\mathbb{R}^{n_0}\rightarrow \mathbb{R}^{n_{\textrm{out}}}$ with $L$ layers. For each layer $l\in [L]$, let $N_l$ denote the set of neurons in layer $l$ and $n_l$ the number of neurons in layer $l$. With the assumption of the piecewise linear activation function, such as the family of ReLU~\citep{montufar2014number,chu2018exact}, the neural network can be represented as a composition of piecewise linear functions $F(\mathbf{x}) = f^{\textrm{out}}\circ f^L\circ f^{L-1}\circ \cdots \circ f^{2}\circ f^{1}(\mathbf{x})$ where $\mathbf{x}\in \mathbb{R}^{n_0}$, $f^l:\mathbb{R}^{n_{l-1}}\rightarrow \mathbb{R}^{n_{l}}$ is a piecewise linear function for $l\in [L]$, and $f^{\textrm{out}}$ is a linear mapping to the final logit. The output of the $l$-th layer is denoted by $\mathbf{x}^l=[\mathbf{x}_1^l,\dots,\mathbf{x}_{n_l}^l]^\top$. Note that the output of the layer refers to the post-activation value of the layer. To express the internal process of the network $F$, we define a function $F^{(i+1):j}(\mathbf{x}^{i})= f^j\circ f^{j-1}\circ \cdots \circ f^{i+2}\circ f^{i+1}(\mathbf{x}^{i})$ as the successive partial representation of $F$, meaning the mapping from layer $i$ to layer $j$. 

\textbf{Decision region.}\; Given instance $\mathbf{x}$, the computation from the intermediate layer to the final logit can be represented as a linear projection
\begin{equation}
    F(\mathbf{x})=
    f^{\textrm{out}}\circ
    f^{(l+1):L}(\mathbf{x}^l)
    =\mathbf{W}\mathbf{x}^l+\mathbf{b}. 
\end{equation}
Note that $\mathbf{W,b}$ depend on $\mathbf{x}$ and $l$. The preimage of $f^{\textrm{out}}\circ f^{(l+1):L}$ is divided into convex polytopes where the function becomes linear for each polytope~\citep{chu2018exact}. We call each polytope as \textit{decision region} since the network applies the same linear projection for the belonging instances to obtain the final logit values. A decision region in the $l$-th layer is determined by the \textit{activated states} of neurons in the higher layers, namely \textit{configuration}.
\begin{definition}[\textbf{Activation State}] 
Given an input $\mathbf{x}\in\mathbb{R}^{n_0}$ and a neuron $i$ in layer $l$ of the neural network, the activation state is
\begin{equation}
    \mathbf{c}_i(\mathbf{x}) = 
    \begin{cases}
0, & \mathrm{\;if}\; \mathbf{x}^{l}_i\leq 0 \\
1, & \mathrm{\;if}\; \mathbf{x}^{l}_i > 0.
    \end{cases}
\end{equation}
\end{definition}
This is the case of the network with ReLU activation. We can easily extend the above definition to the other piecewise linear activation functions. The activation states can be represented as a vector code with discrete values. We define this code as a configuration.
\begin{definition}[\textbf{Configuration}]
Given an input $\mathbf{x}\in\mathbb{R}^{n_0}$ and a set of neurons $N$, the configuration is
\begin{equation}
    \mathbf{c}^N(\mathbf{x})=[\mathbf{c}_{N[1]}(\mathbf{x}),\dots,\mathbf{c}_{N[|N|]}(\mathbf{x})]
\end{equation} 
where $N[i]$ denotes the $i$-th neuron in set $N$.
\end{definition}
In consequence, a decision region where the given $\mathbf{x}$ located in the $l$-th layer is determined by $\mathbf{c}(\mathbf{x})=\mathrm{concat}([\mathbf{c}^{N_{l+1}}(\mathbf{x}),\dots,\mathbf{c}^{N_L}(\mathbf{x})])$. Unless we specify the neuron set $N$, a configuration of $\mathbf{x}$ denotes $\mathbf{c(x)}$, considering every neurons in higher layers. 

\textbf{Internal decision boundary.}\; We define an \textit{internal decision boundary} as the boundary within the feature space where a transition in the activation state of each neuron occurs. In other words, each element of $\mathbf{c}(\mathbf{x})$ implies which state $\mathbf{x}$ is located with respect to the corresponding internal decision boundary. Note that the term decision boundary typically refers to the boundary where the classification results change in other literature. However, in this paper, we use the term decision boundary as the internal decision boundary. 

\section{Methods}
\label{sec:featurespace}

The configuration, representing the activated states of the internal decision boundaries, determines the decision region to which an instance pertains at the target layer. This region serves as a guide for the model to extract pivotal information from the feature space. From the perspective of distributed representations, this information is captured by a specific subset of neurons, as experimentally observed by~\citet{fong2018net2vec}. In this spirit, to understand the nature of the captured information for a target instance, it is imperative to identify a subset of principal neurons shared with relevant instances.

Following this logic, we present an interpretation framework designed to identify an interpretable region that aligns with the desired concept properties of learned concepts: learned concepts with coherence and discrimination. Our method automatically finds a set of instances sharing a concept of target instance through \textit{Configuration distance}, and forms a \textit{Relaxed Decision Region} by extracting principal neurons that represent this concept.

\subsection{Configuration Distance}
Before identifying a subset of principal neurons, the initial step involves finding the concept set. In contrast to other methods that heavily rely on pre-defined concept sets, we automatically discover a group of instances that share learned concepts with a given target instance. To enable this process, we introduce a metric to evaluate the difference in configurations as follows.

\begin{definition}[\textbf{Configuration Distance}] \;
Given an instance $\mathbf{x}, \tilde{\mathbf{x}} \in\mathcal{X}$, the Configuration distance for a set of neurons $N$ is defined as follows:
\begin{equation}
\label{eq:Config-dist}
d_{C}(\mathbf{x}, \tilde{\mathbf{x}}) = d_H(\mathbf{c}^N(\mathbf{x}),\mathbf{c}^N(\tilde{\mathbf{x}}))
\end{equation}
where $d_H$ denotes the Hamming distance.
\end{definition}

$N$ can be selected from either a single layer or multiple layers. If we want to focus on the specific projection from the $l$-th layer to the (\textit{l}+1)-th layer, the configuration at the (\textit{l}+1)-th layer, denoted as $d^{N_l}_C$, would be considered. We empirically verify in the following sections that concept sets are effectively found through the Configuration distance.

\subsection{Relaxed Decision Region}
\label{sec:RDR}

From the configuration of a target, we select a principal subset of neurons. The activation states of these neurons construct an integrated decision region in the feature space where encompassed instances share the learned concept. We call the states of these selected neurons as the \textit{principal configuration} $\mathbf{c}_{\mathrm{p}}$ and the corresponding region as a \textit{Relaxed Decision Region (RDR)}, $\mathcal{R}$. Finding a principal configuration can be formulated as follows:
\begin{equation}
\label{opt:pc}
\begin{aligned}
    \min_{\underset{\scriptstyle N^*\subset N}{\mathbf{c}_{\mathrm{p}}\in \{0,1\}^t}}
     &\;\mathbb{E}_{\mathbf{x}}[d_H(\mathbf{c}^{N^*}(\mathbf{x}),\mathbf{c}_{\mathrm{p}})] - 
    \mathbb{E}_{\mathbf{y}}[d_H(\mathbf{c}^{N^*}(\mathbf{y}),\mathbf{c}_{\mathrm{p}})]\\
    \mathrm{s.t.}\;\;  &|N^{*}|=t 
\end{aligned}
\end{equation}
where $\mathbf{x}$, $\mathbf{y}$ represent random variables corresponding to the positive (concept) set of inputs and the negative (concept) set of inputs, respectively. We find a principal subset $N^*$ that consists of $t$ number of neurons from a neuron set $N$. Minimizing the objective function encourages the principal configuration to exhibit strong coherence with the positive set (the first term) while also ensuring distinctiveness from the negative set (the second term). 

In practice, we construct a positive set $S$ and a negative set $S_{\mathrm{neg}}$ from training data. For a given target instance, we collect $k$-nearest neighbors (including itself) based on the $d_C^N$ and assign them to the positive set $S$. The negative set $S_{\mathrm{neg}}$ can be easily set to the remaining data points. One of the strengths of our framework here is that it does not require a pre-defined concept set. To address the optimization problem in Equation (\ref{opt:pc}), we employ a greedy algorithm for assigning neurons to $N^*$. Our greedy algorithm is described in Algorithm~\ref{alg:RDR}.

\begin{theorem}
The optimal solution $N^*$ of the problem in Equation (\ref{opt:pc}) can be obtained by the greedy algorithm.
\end{theorem}
\begin{proof}
    See Appendix.
\end{proof}

\begin{algorithm}[ht]
\caption{Finding a Relaxed Decision Region}
\label{alg:RDR}
\begin{flushleft}
        \textbf{Input:}  $S$, $S_{\textrm{neg}}$, a set of neurons $N$, layer $l$\\
        \textbf{Parameter:} the number of neurons to select $t$\\
        \textbf{Output:} $N^*$, $\mathbf{c}_{\mathrm{p}}$, $\mathcal{R}$
\end{flushleft}
\begin{algorithmic}[1]
\STATE Initialize $N^*$=\{\}, $\mathbf{c}_{\mathrm{p}}\in\{0,1\}^{t}$
\STATE $\bar{\textbf{c}}=\frac{1}{|S|} \sum_{\mathbf{x}\in S}
        \mathbf{c}^N(\mathbf{x})$
\STATE$\bar{\textbf{c}}_{\mathrm{neg}}=\frac{1}{|S_{\mathrm{neg}}|} \sum_{\mathbf{y}\in S_{\mathrm{neg}}}
        \mathbf{c}^N(\mathbf{y})$
   \FOR{$i=1,\dots,t$}
        \STATE $i^* = \mathrm{argmax}_{i \in N}|\bar{\mathbf{c}}_i-\bar{\mathbf{c}}_{\mathrm{neg},i}|$
        \STATE $\mathbf{c}_{\mathrm{p},i^*}=\bar{\mathbf{c}}_{i^*}$
        \STATE $N^*=N^*\cup \{i^*\}$ and $N=N\setminus \{i^*\}$
   \ENDFOR
   \STATE $\mathcal{R} = \{ \mathbf{x}^l \; |\; 
   d_H(\mathbf{c}^{N^*}(\mathbf{x}),\mathbf{c}_{\mathrm{p}})=0, \mathbf{x}\in \mathcal{X} \} $
\end{algorithmic}
\end{algorithm}
In our problem, the candidate neuron set $N$ consists of neurons that exhibit identical activation states for all instances in $S$ so that RDR encompasses all instances in $S$. $\bar{\textbf{c}}$, $\bar{\textbf{c}}_{\mathrm{neg}}$ denote the frequencies of neuron activations for instances in $S$ and $S_\mathrm{neg}$, respectively. Our greedy algorithm then iteratively selects a neuron that has the largest frequency difference between $S$ and $S_\mathrm{neg}$, indicating a high information gain to explain $S$. As each neuron is associated with an internal decision boundary in the feature space, we select a subset of the boundaries to form a larger region for coherent interpretation. This is why we use the term `Relaxed’ in RDR.

The parameter $t$ controls the number of principal neurons. A smaller $t$ makes the RDR looser so that it captures a more general concept while a larger $t$ leads to detecting more specific properties. The number of neighbors $k$ gives a similar but opposite effect to the results with $t$. To mitigate the concern about the difficulty of parameter selection, we provide guidance on selecting $t$ and $k$ in Appendix. We empirically check that the RDR works effectively with the parameters $k\in[5,10]$ and $t\in[9,15]$ in the penultimate convolutional block of the models in our experiments.

\textbf{Geometric understanding of RDR.}\; Figure~\ref{fig:main-concept} illustrates how RDR captures instances with learned concepts in the feature space from the geometric view. We selected three instances, two of which seem similar while the last one has distinct semantics. Then, we drew the 2-d plane that passes the feature maps of three given instances on the feature space of the 12th layer in VGG19. Each line on the plane represents an internal decision boundary corresponding to each neuron in the higher layers. Although the third instance has a smaller Euclidean distance value, the second instance (with a smaller Configuration distance value) is much more akin to the first instance. Indeed, We can easily check that there are numerous internal decision boundaries between the third instance and the others. 

This intricate space partition enables DNN to apply different mappings according to inputs~\cite{chu2018exact, gopinath2019property}. Our RDR framework finds core internal decision boundaries and relaxes the decision regions where mappings are similar (highlighted in green in Figure~\ref{fig:main-concept}). Further explanations are provided in Appendix.

\section{Analysis for the Distance metrics}

We demonstrate the effectiveness of the Configuration distance in disclosing learned concepts in the feature space compared to other standard metrics. Our analysis gives insights into the characteristics of decision regions. 

We compare the Configuration distance with two standard distance metrics, the Euclidean distance and the Cosine distance, in terms of conceptual similarity among the nearest instances. The Configuration distance effectively captures instances with similar concepts in the feature space, whereas the Euclidean distance tends to fail in evaluating resemblance. In the case of the Cosine distance, while the closest neighbors usually include appropriate instances, some less relevant instances are also present among the neighbors.

\begin{figure}[ht!]
\centering
\includegraphics[width=\columnwidth]{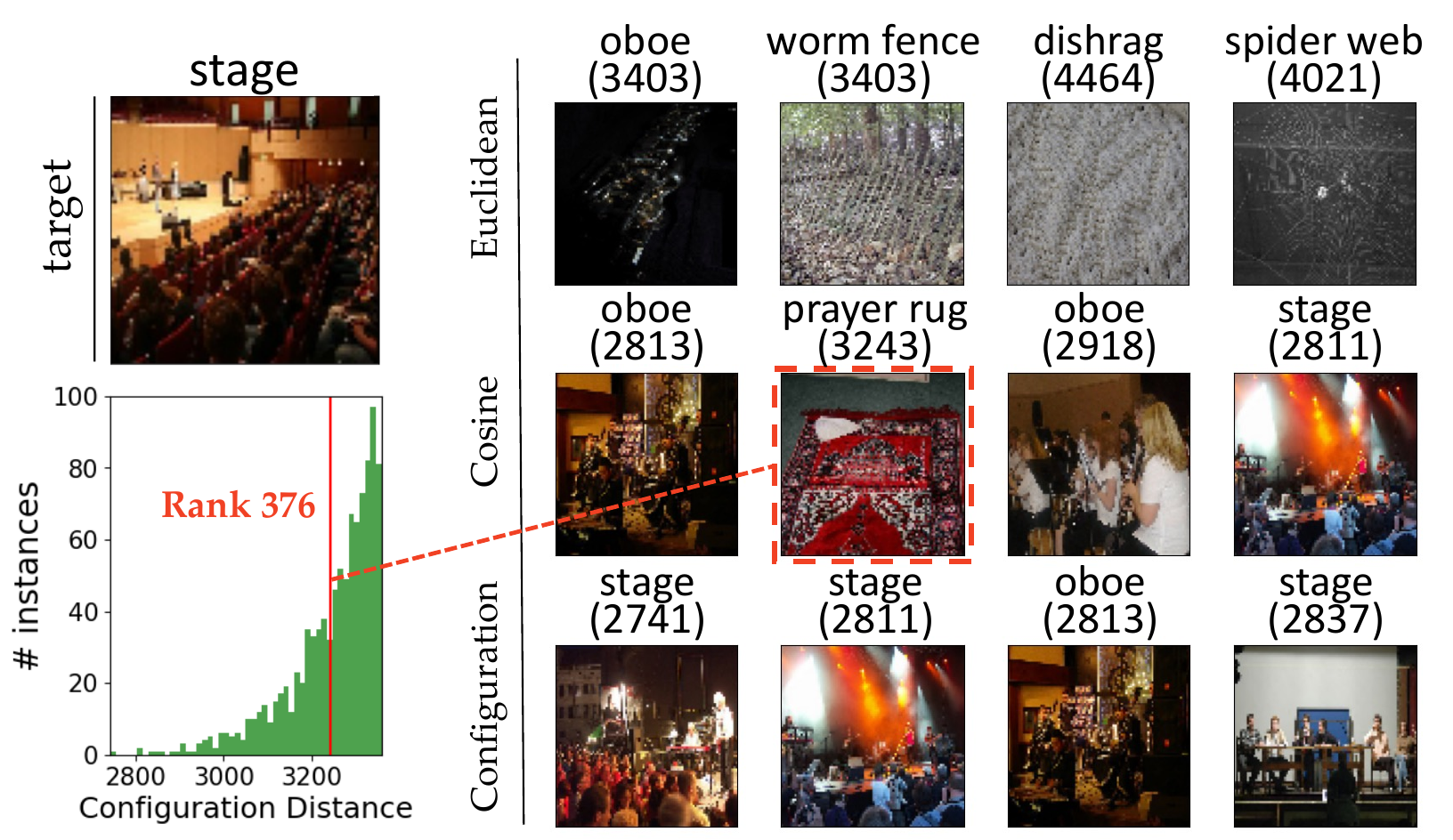}
\caption{\normalsize Top 4 nearest instances with different distance metrics. In the case of the Euclidean distance and the Cosine distance, the irrelevant instances are detected. These instances have large Configuration distances from the target.}
\label{fig:metric-comp}
\end{figure}
\normalsize
In Figure~\ref{fig:metric-comp}, we visualize the top 4 nearest instances of the target `stage' image, based on each distance metric. The number in parentheses denotes the Configuration distance to the target. It helps to successfully detect `people in the stage-like place', while we cannot attain meaningful information from the Euclidean distance. For the Cosine distance, the second image, which is semantically distinct, is far away from the target with respect to the Configuration distance. The histogram shows the distribution of the 1000 smallest Configuration distance values within the training data. Among these, 375 instances are closer to the target than the `prayer rug' image, and none of the top 4 Euclidean images are included. This serves as compelling evidence of the effectiveness of the Configuration distance in measuring differences in learned concepts.

\textbf{Insight.} \; We observe that for a given target instances with smaller Cosine distances tend to have smaller Configuration distances, which supports the efficacy of Cosine distance in evaluating similarity in the intermediate feature space compared to the Euclidean distance (more examples are in Appendix). We conjecture that this phenomenon is attributed to the geometry created by DNN structures. For example, in CNNs, decision regions are divided by polyhedral cones~\cite{carlsson2019geometry} so that the angular difference between feature maps becomes highly related to the Configuration distance. This aligns with the empirical successes of prior work using the Cosine similarity in the feature space~\cite{fong2018net2vec, kim2018interpretability, bachman2019learning, jeon2020egbas}. We plan to explore this phenomenon further in our future work.

\begin{figure}[ht!]
\centering
\includegraphics[width=\columnwidth]{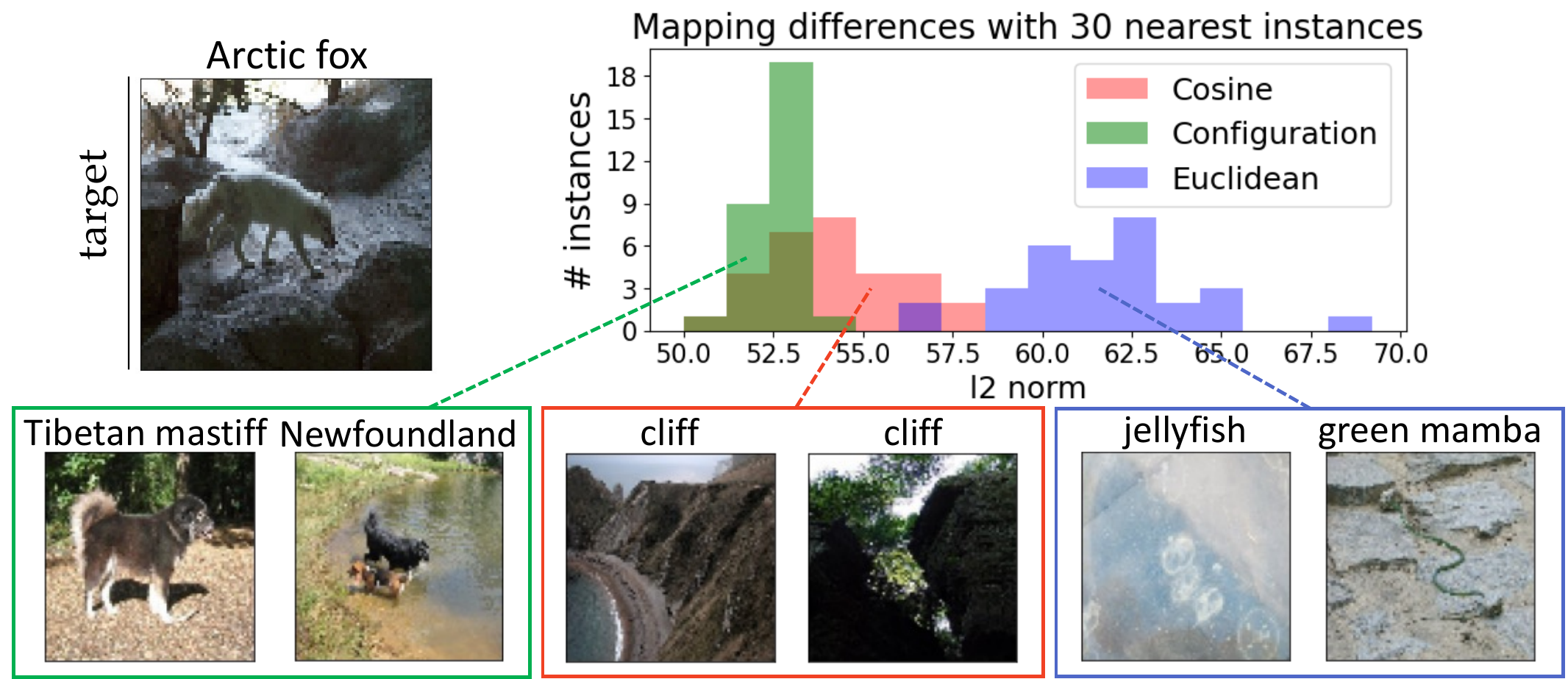}
\caption{\normalsize Mapping differences with 30 nearest neighbors. The Configuration distance indeed captures instances whose mappings are close to the target's one. With a smaller mapping difference, the image is more similar to the target.}
\label{fig:weight-diff}
\end{figure}

\textbf{The viewpoint of Mapping.} \; So far, we have explained that DNNs extract different information from instances due to changes in mapping according to the configuration. In Figure~\ref{fig:weight-diff}, we compute the difference in mappings of 30 nearest instances. The difference is quantified using the L2 norm of the weight matrices in two successive layers. Compared to the other distance metrics, the Configuration distance captures instances whose mappings are close to the target's one, leading to the extraction of more similar information.

\section{Experiments}
\label{sec:exp}

In this section, we present the qualitative and quantitative evaluation results of our proposed method as well as various use cases. Our experiments are conducted on the Mini-ImageNet \citep{vinyals2016matching}, Flowers Recognition (denoted by Flowers), Oxford pet, Broden~\citep{bau2017network}, Imagenet-X~\citep{idrissi2022imagenet} datasets, using VGG19~\citep{simonyan2014very}, ResNet50~\citep{he2016deep}, and MobileNetV2~\citep{sandler2018mobilenetv2} models. The choice of parameters $t$ and $k$ adheres to the criteria outlined in the previous section unless explicitly stated. Detailed settings of each experiment are provided in Appendix. 

\subsection{Coherence of Captured Concepts} 
\label{exp:heatmap}

\begin{figure}[ht]
\centering
\includegraphics[width=\columnwidth]{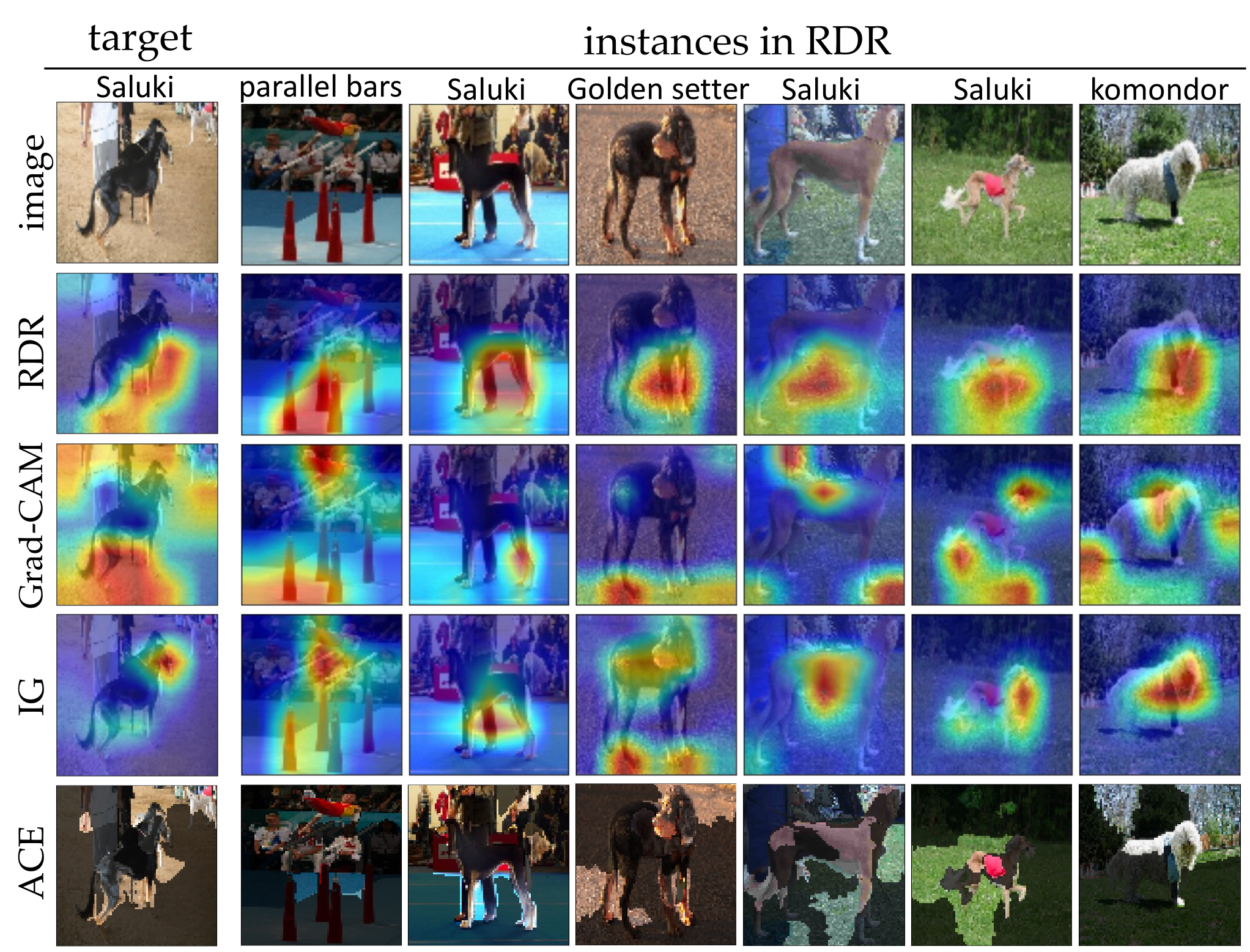}
\caption{\normalsize Retrieval results using RDR and masking for Concept Location. Without supervision, RDR successfully groups similar instances and related parts. The concept of `legs' is learned for the images in the 12th layer.}
\label{fig:masking}
\end{figure}

To assess the coherence of captured concepts, we identify which parts in the image correspond to concepts with principal configurations from a convolutional layer. Following the methods of~\citet{bau2017network} and \citet{fong2018net2vec}, we visualize activation maps of the channels that contain the neurons in the principal configuration.

Figure~\ref{fig:masking} shows our visualization results compared to those from other interpretability methods: Grad-CAM~\citep{selvaraju2017grad}, IG~\citep{sundararajan2017axiomatic} and ACE~\citep{ghorbani2019towards}. To apply IG for the intermediate layer, we compute the attribution scores for each neuron in the feature map and sum the scores across channels at each spatial location. In ACE, we adhere to the settings outlined in the original paper and visualize images by masking all except the top 30$\%$ of significant segmentation patches. These results are obtained from the 12th layer in VGG19 using $d_C^{12}$ with $k$=8, $t$=10. The instances in RDR share the concept of `legs' and the selected channels focus on the `legs' parts. Grad-CAM fails to find appropriate parts in the middle layer. While IG emphasizes specific parts of an object, it lacks consistency across the images. ACE prioritizes features for classification rather than maintaining a coherent conceptual interpretation. These results support the necessity of group-level interpretation to understand distributed representations, moving beyond the consideration of class-specific information such as gradients.


\begin{figure}[ht!]
\centering
\centerline{\includegraphics[width=\columnwidth]{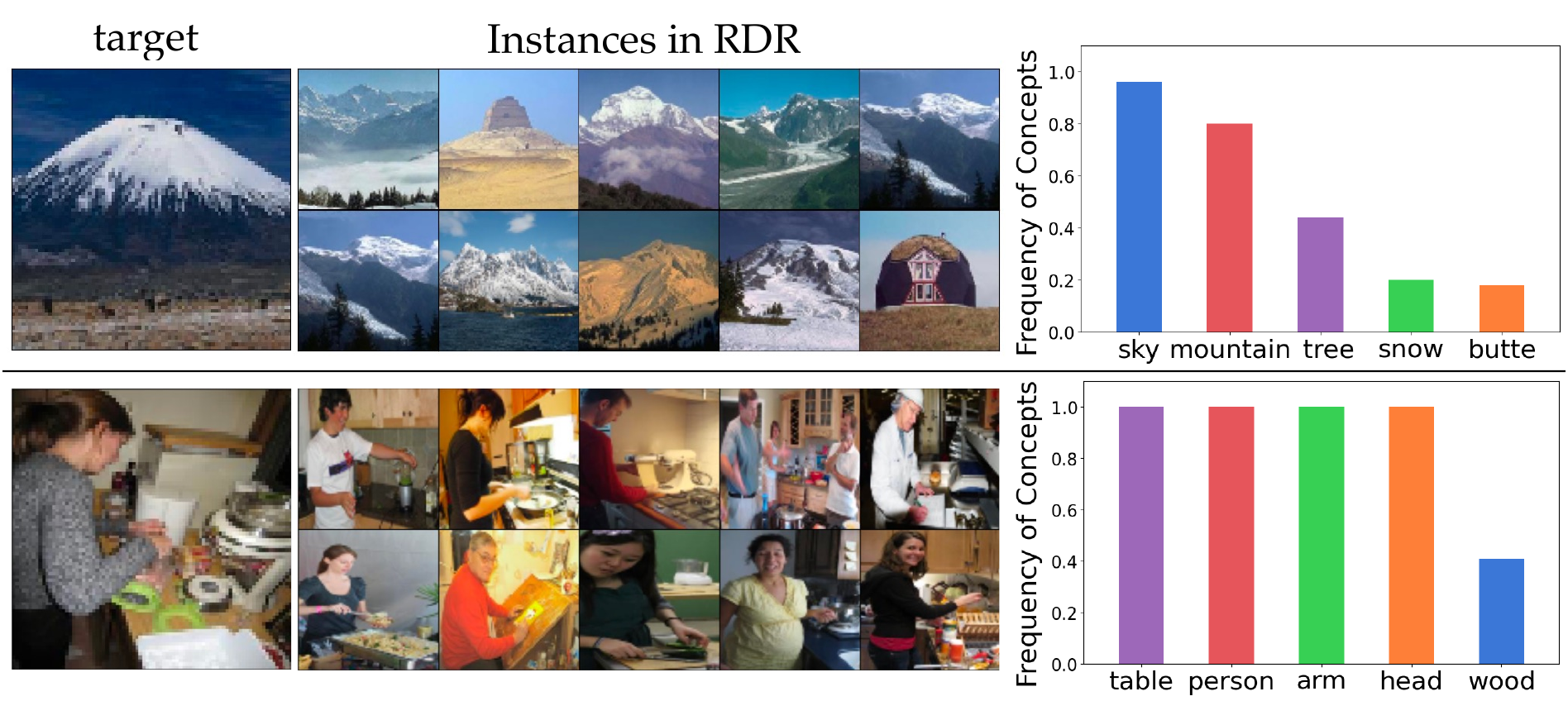}}
\caption{\normalsize Identifying the coherent properties using human-annotated information in the Broden dataset.}
\label{fig:broden}
\end{figure}

We conduct additional investigations to assess whether concepts identified across instances in RDR align with pre-defined concept labels in the Broden dataset. The Broden dataset provides pixel-wise segmentation concept labels for each image. In Figure~\ref{fig:broden}, we initially collect instances within an RDR (left) and identify the top 5 concept labels that they share (right). The experiment follows the same setting in Figure~\ref{fig:masking} but with $k$=10, $t$=15. The experimental results clearly demonstrate that the concepts captured with an RDR align well with human-labeled concepts.

\subsection{Identifying Learned Concepts over Layers}
\label{sec:layers}
\begin{figure}[ht] 
\centering
\centerline{\includegraphics[width=\columnwidth]{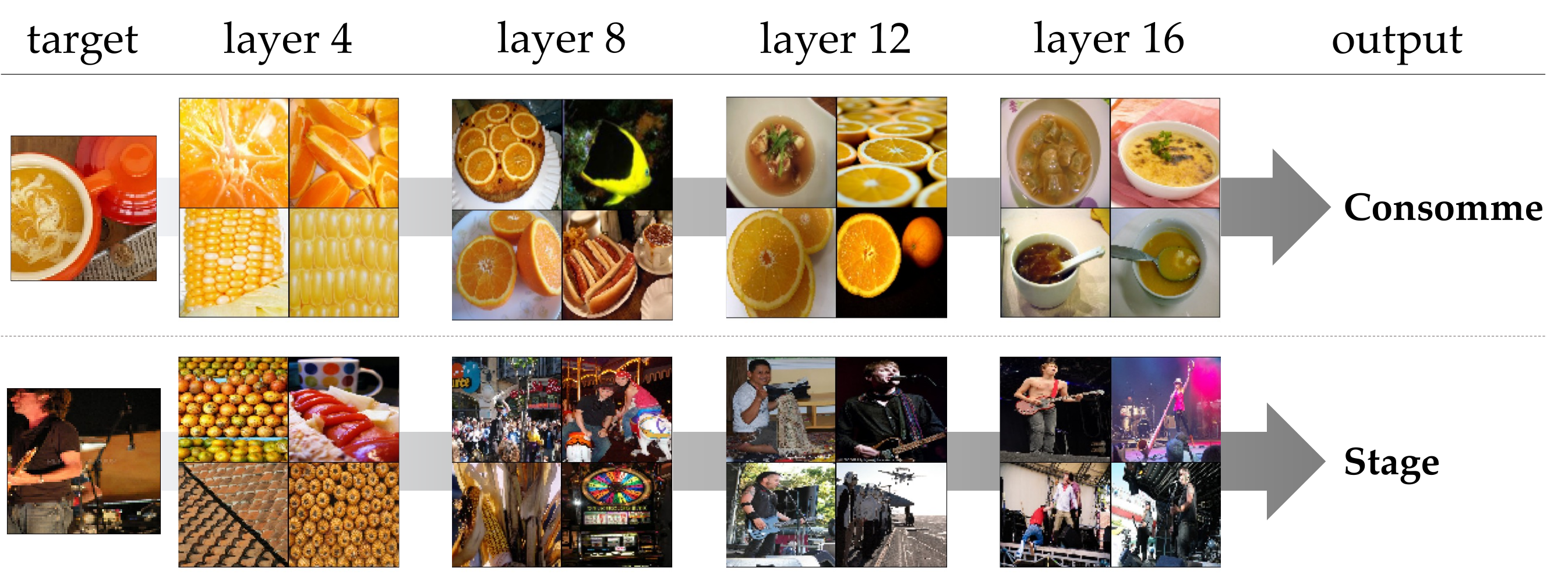}}
\caption{\normalsize Differences in RDR over layers. The higher the layer is, the more class-specific concepts are captured.} 
\label{fig:layers}
\end{figure}

By constructing RDR across various layers, we investigate how DNN recognizes instances at different stages of its architecture. Figure~\ref{fig:layers} illustrates the layer-wise differences of RDRs in the feature space of VGG19. We observe that the lower layers tend to capture spatial information, such as the object shape, whereas higher layers learn more detailed and class-specific features. The results align with the observations identified by~\citet{bau2017network}. Additional examples are provided in Appendix.

\subsection{Reasoning Misclassified Cases} 
\label{sec:misclass}

\begin{figure}[ht] 
\centering
\centerline{\includegraphics[width=\columnwidth]{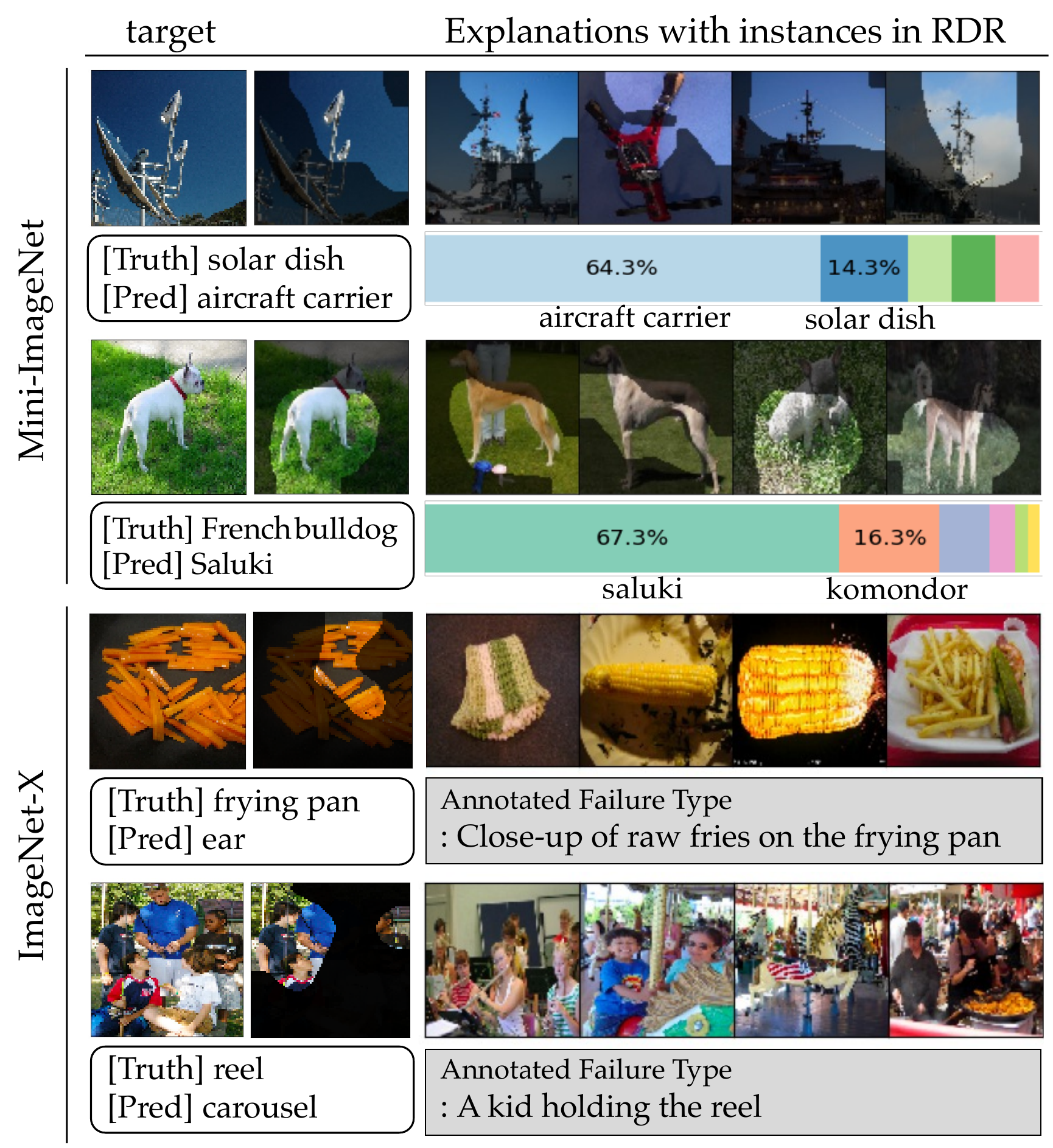}}
\caption{\normalsize Misclassification reasoning can be investigated by examining RDR. In the Mini-ImageNet case, we provide the ratio of classes in RDR. In the ImageNet-X case, we describe the annotated failure type.} 
\label{fig:misclass}
\end{figure}

We leverage RDR to comprehend the causes of misclassifications, under the assumption that certain internal neurons encode spurious correlations with actual labels, leading to classification errors. In Figure~\ref{fig:misclass}, we present which parts in the image contribute to misclassifications by designating $S_{\mathrm{neg}}$ as instances with the true label of the target. By visualizing the channels associated with the principal configuration, we can obtain evidence for each misclassification case. For instance, the second row demonstrates that the misclassification of the target instance stems from the presence of long legs, a characteristic commonly linked with Saluki. To further validate our findings, we double-check whether the extracted mislearned concepts align with human-labeled failure reasons from ImageNet-X. Additional illustrative examples are presented in the Appendix.

\subsection{Finding Unlabeled Subclasses} 
Discovering unlabeled subclasses without human supervision is a challenging task. Our framework can reveal subclasses inherent in data without any prior knowledge. In Figure~\ref{fig:main-result}, we detect subclasses in the Mini-ImageNet dataset and the Flowers dataset captured by VGG19. We compute the Configuration distance at target layers $\{12,14,16,17\}$. For each class, three target instances are chosen, followed by displaying 10 randomly selected images from the corresponding RDRs. Consequently, RDRs successfully capture learned concepts, including subclasses, in an unsupervised way. This qualitative evidence supports why our framework achieves good performance on the quantitative comparison shown in Table~\ref{table1}.

\begin{table}
\centering
\fontsize{9}{11}\selectfont
\begin{tabularx}{\columnwidth}{c|ccc|ccc}
\noalign{\smallskip}\noalign{\smallskip}\hline\hline
\multirow{2}{*} & \multicolumn{3}{c|}{Purity} & \multicolumn{3}{c}{Entropy}\\ 
\hline
& VGG & RSN & MBN  & VGG  & RSN  & MBN \\ 
\hline
\textbf{RDR}  & \textbf{0.351} & \textbf{0.408} & \textbf{0.346}  & \textbf{1.527} & \textbf{1.372} & 1.531  \\
$KNN_{C}$  & 0.303  & 0.328   & 0.329 & 1.588   & 1.497 & \textbf{1.498} \\
$CAR_{C}$ & 0.022 & 0.038 & 0.036 & 2.264 & 2.153 & 2.527 \\
$CAV_{C}$ & 0.314 & 0.387 & 0.323 & 1.549 & 1.416 & 1.575\\
RI  & 0.045 & 0.056 & 0.056  & 2.161 & 1.971  & 2.369   \\
\hline
$RDR_{Euc}$ & 0.241 & 0.252 & 0.303 & 1.76 & 1.779 & 1.76 \\
$KNN_{Euc}$  & 0.183 & 0.166 & 0.275 & 1.835  & 1.862 & 1.791  \\
$CAR_{Euc}$ & 0.039 & 0.037 & 0.037 & 2.272 & 2.17 & 2.476 \\
$CAV_{Euc}$ & 0.207 & 0.240 & 0.283 & 1.811 & 1.787 & 1.745 \\
$RDR_{Cos}$ & 0.309 & 0.307 & 0.346 & 1.613 & 1.7 & 1.628 \\
$KNN_{Cos}$ & 0.250 & 0.232 & 0.283 & 1.672 & 1.771 & 1.635 \\
$CAR_{Cos}$ & 0.042 & 0.027 & 0.036 & 2.251 & 2.14 & 2.576 \\
$CAV_{Cos}$ & 0.261 & 0.283 & 0.274 & 1.596 & 1.734 & 1.651\\
\hline 
\hline 
\end{tabularx}
\caption{\normalsize Quantitative results for evaluating the coherence of subclass grouping. VGG, RSN, MBN represent VGG19, ResNet50, and MobileNetV2, respectively.} 
\label{table1}
\end{table}

\subsection{Quantitative Evaluation} 

We quantitatively evaluate the coherence of subclass groups identified by RDR, comparing with other methods on the Oxford-pet dataset. Although each model is trained to solely distinguish between cat and dog images, our RDR framework can implicitly identify specific breeds (subclasses) of target instances, even in the absence of explicit breed information. For a thorough verification, we form RDRs with 50 randomly chosen target instances, ensuring an equal number of instances within each group across all methods.

In Table~\ref{table1}, we employ two metrics, namely \textit{purity} and \textit{entropy}, for comprehensive quantitative evaluations~\citep{zhao2001criterion}. Purity assesses the proportion of samples containing the target subclass within the group. Entropy measures the uncertainty of subclasses within the group. $$\textrm{Purity} = \frac{1}{T} \sum_{t=1}^T \mathbf{1}_{[y_t = \tilde{y}]}$$ 
$$\textrm{Entropy} = \sum_{y:\mathcal{P}_y \neq 0} -\mathcal{P}_y* \log \mathcal{P}_y$$ where $\mathcal{P}_y = \frac{1}{T} \sum_{t=1}^T \mathbf{1}_{[y_t= y]}$ (empirical distribution). $T$ is the number of samples in each group, and $y_t$ is the subclass of $t$-th sample in a group. Subclass is denoted as $y$ and that of a target instance is $\tilde{y}$. High purity or low entropy indicates that the group consistently extracts the subclass of the target.

We compare RDR with other approaches that can define interpretable regions: K-Nearest Neighborhood, Representative Interpretation \citep{lam2021finding}, CAR \citep{crabbé2022concept} and CAV \citep{kim2018interpretability}. To ensure a fair comparison, we consider the nearest neighbors as a concept set for CAR and CAV, as they require pre-defined concept sets. In the case of CAV, since the original CAV does not output a region, we define a region for CAV by containing instances with have high cosine similarities to the computed CAV. As shown in Table~\ref{table1}, RDR generally outperforms others in both purity and entropy.

Following its original settings, CAR exhibits an excessively broad region, grouping various subclasses into a single region. Similarly, RI also generates a wide concept region, as it is formulated to maximize the number of samples in a target class, potentially overlooking implicit concepts. The competitive results observed in CAV can be attributed to the phenomenon where a high cosine similarity may result in a low Configuration distance, as elucidated in our analysis section. We also ablate the methods using different distance metrics, confirming the effectiveness of configuration. 

\section{Conclusion}

We introduce a novel interpretation framework that reveals the learned concepts in DNNs without human supervision. Our key approach is to leverage the activation states to identify the distributed representations of concepts. We propose the Configuration Distance, a novel metric that effectively assesses the disparity in decision regions. It enables the automatic collection of concept sets, avoiding the need for pre-defined information. By extracting the principal configuration from the set, we construct a Relaxed Decision Region (RDR) that provides consistent interpretation for the related instances. In our experiments, we present various applications of RDR for interpreting DNNs, including subclass detection, reasoning misclassification, and exploring layer-wise concepts. We expect that our work guides the direction to understanding the decision-making process of DNNs, which is a crucial step for real-world applications.

\section*{Acknowledgements}
This work was partly supported by Institute of Information \& Communications Technology Planning \& Evaluation (IITP) grant funded by the Korea government (MSIT) (No. 2022-0-00984, Development of Artificial Intelligence Technology for Personalized Plug-and-Play Explanation and Verification of Explanation, 50\%; No. 2022-0-00184, Development and Study of AI Technologies to Inexpensively Conform to Evolving Policy on Ethics, 30\%; No. 2021-0-02068, Artificial Intelligence Innovation Hub, 10\%; No. 2019-0-00075, Artificial Intelligence Graduate School Program (KAIST), 10\%).

\bibliography{aaai24}

\begin{thebibliography}{41}
\providecommand{\natexlab}[1]{#1}

\bibitem[{Achtibat et~al.(2022)Achtibat, Dreyer, Eisenbraun, Bosse, Wiegand, Samek, and Lapuschkin}]{achtibat2022towards}
Achtibat, R.; Dreyer, M.; Eisenbraun, I.; Bosse, S.; Wiegand, T.; Samek, W.; and Lapuschkin, S. 2022.
\newblock From" where" to" what": Towards human-understandable explanations through concept relevance propagation.
\newblock \emph{arXiv preprint arXiv:2206.03208}.

\bibitem[{Angelov(2020)}]{angelov2020top2vec}
Angelov, D. 2020.
\newblock Top2Vec: Distributed Representations of Topics.
\newblock arXiv:2008.09470.

\bibitem[{Bach et~al.(2015)Bach, Binder, Montavon, Klauschen, M{\"u}ller, and Samek}]{bach2015pixel}
Bach, S.; Binder, A.; Montavon, G.; Klauschen, F.; M{\"u}ller, K.-R.; and Samek, W. 2015.
\newblock On pixel-wise explanations for non-linear classifier decisions by layer-wise relevance propagation.
\newblock \emph{PloS one}, 10(7): e0130140.

\bibitem[{Bachman, Hjelm, and Buchwalter(2019)}]{bachman2019learning}
Bachman, P.; Hjelm, R.~D.; and Buchwalter, W. 2019.
\newblock Learning representations by maximizing mutual information across views.
\newblock \emph{Advances in neural information processing systems}, 32.

\bibitem[{Bau et~al.(2017)Bau, Zhou, Khosla, Oliva, and Torralba}]{bau2017network}
Bau, D.; Zhou, B.; Khosla, A.; Oliva, A.; and Torralba, A. 2017.
\newblock Network dissection: Quantifying interpretability of deep visual representations.
\newblock In \emph{Proceedings of the IEEE conference on Computer Vision and Pattern Recognition}, 6541--6549.

\bibitem[{Carlsson(2019)}]{carlsson2019geometry}
Carlsson, S. 2019.
\newblock Geometry of deep convolutional networks.
\newblock \emph{arXiv preprint arXiv:1905.08922}.

\bibitem[{Chattopadhay et~al.(2018)Chattopadhay, Sarkar, Howlader, and Balasubramanian}]{chattopadhay2018grad}
Chattopadhay, A.; Sarkar, A.; Howlader, P.; and Balasubramanian, V.~N. 2018.
\newblock Grad-cam++: Generalized gradient-based visual explanations for deep convolutional networks.
\newblock In \emph{2018 IEEE winter conference on applications of computer vision (WACV)}, 839--847. IEEE.

\bibitem[{Chen et~al.(2019)Chen, Li, Tao, Barnett, Rudin, and Su}]{chen2019looks}
Chen, C.; Li, O.; Tao, D.; Barnett, A.; Rudin, C.; and Su, J.~K. 2019.
\newblock This looks like that: deep learning for interpretable image recognition.
\newblock \emph{Advances in neural information processing systems}, 32.

\bibitem[{Cho et~al.(2021)Cho, Chang, Lee, and Choi}]{cho2021interpreting}
Cho, S.; Chang, W.; Lee, G.; and Choi, J. 2021.
\newblock Interpreting internal activation patterns in deep temporal neural networks by finding prototypes.
\newblock In \emph{Proceedings of the 27th ACM SIGKDD Conference on Knowledge Discovery \& Data Mining}, 158--166.

\bibitem[{Chu et~al.(2018)Chu, Hu, Hu, Wang, and Pei}]{chu2018exact}
Chu, L.; Hu, X.; Hu, J.; Wang, L.; and Pei, J. 2018.
\newblock Exact and consistent interpretation for piecewise linear neural networks: A closed form solution.
\newblock In \emph{Proceedings of the 24th ACM SIGKDD International Conference on Knowledge Discovery \& Data Mining}, 1244--1253.

\bibitem[{Crabbé and van~der Schaar(2022)}]{crabbé2022concept}
Crabbé, J.; and van~der Schaar, M. 2022.
\newblock Concept Activation Regions: A Generalized Framework For Concept-Based Explanations.
\newblock arXiv:2209.11222.

\bibitem[{Fong and Vedaldi(2018)}]{fong2018net2vec}
Fong, R.; and Vedaldi, A. 2018.
\newblock Net2vec: Quantifying and explaining how concepts are encoded by filters in deep neural networks.
\newblock In \emph{Proceedings of the IEEE conference on computer vision and pattern recognition}, 8730--8738.

\bibitem[{Ghorbani et~al.(2019)Ghorbani, Wexler, Zou, and Kim}]{ghorbani2019towards}
Ghorbani, A.; Wexler, J.; Zou, J.~Y.; and Kim, B. 2019.
\newblock Towards automatic concept-based explanations.
\newblock \emph{In Proceedings of the Advances in Neural Information Processing Systems}, 32.

\bibitem[{Gopinath et~al.(2019)Gopinath, Converse, Pasareanu, and Taly}]{gopinath2019property}
Gopinath, D.; Converse, H.; Pasareanu, C.; and Taly, A. 2019.
\newblock Property inference for deep neural networks.
\newblock In \emph{2019 34th IEEE/ACM International Conference on Automated Software Engineering (ASE)}, 797--809. IEEE.

\bibitem[{Gunning et~al.(2019)Gunning, Stefik, Choi, Miller, Stumpf, and Yang}]{gunning2019xai}
Gunning, D.; Stefik, M.; Choi, J.; Miller, T.; Stumpf, S.; and Yang, G.-Z. 2019.
\newblock XAI—Explainable artificial intelligence.
\newblock \emph{Science robotics}, 4(37): eaay7120.

\bibitem[{He et~al.(2016)He, Zhang, Ren, and Sun}]{he2016deep}
He, K.; Zhang, X.; Ren, S.; and Sun, J. 2016.
\newblock Deep residual learning for image recognition.
\newblock In \emph{Proceedings of the IEEE conference on computer vision and pattern recognition}, 770--778.

\bibitem[{Hinton(1984)}]{hinton1984distributed}
Hinton, G.~E. 1984.
\newblock Distributed representations.

\bibitem[{Hinton et~al.(1986)}]{hinton1986learning}
Hinton, G.~E.; et~al. 1986.
\newblock Learning distributed representations of concepts.
\newblock In \emph{Proceedings of the eighth annual conference of the cognitive science society}, volume~1, 12. Amherst, MA.

\bibitem[{Idrissi et~al.(2022)Idrissi, Bouchacourt, Balestriero, Evtimov, Hazirbas, Ballas, Vincent, Drozdzal, Lopez-Paz, and Ibrahim}]{idrissi2022imagenet}
Idrissi, B.~Y.; Bouchacourt, D.; Balestriero, R.; Evtimov, I.; Hazirbas, C.; Ballas, N.; Vincent, P.; Drozdzal, M.; Lopez-Paz, D.; and Ibrahim, M. 2022.
\newblock Imagenet-x: Understanding model mistakes with factor of variation annotations.
\newblock \emph{arXiv preprint arXiv:2211.01866}.

\bibitem[{Jeon, Jeong, and Choi(2020)}]{jeon2020egbas}
Jeon, G.; Jeong, H.; and Choi, J. 2020.
\newblock An Efficient Explorative Sampling Considering the Generative Boundaries of Deep Generative Neural Networks.
\newblock \emph{Proceedings of the AAAI Conference on Artificial Intelligence}, 34(04): 4288--4295.

\bibitem[{Khanna et~al.(2019)Khanna, Kim, Ghosh, and Koyejo}]{khanna2019interpreting}
Khanna, R.; Kim, B.; Ghosh, J.; and Koyejo, S. 2019.
\newblock Interpreting black box predictions using fisher kernels.
\newblock In \emph{The 22nd International Conference on Artificial Intelligence and Statistics}, 3382--3390.

\bibitem[{Kim, Khanna, and Koyejo(2016)}]{kim2016examples}
Kim, B.; Khanna, R.; and Koyejo, O.~O. 2016.
\newblock Examples are not enough, learn to criticize! criticism for interpretability.
\newblock \emph{Proceedings of the Advances in Neural Information Processing Systems}, 29.

\bibitem[{Kim et~al.(2018)Kim, Wattenberg, Gilmer, Cai, Wexler, Viegas et~al.}]{kim2018interpretability}
Kim, B.; Wattenberg, M.; Gilmer, J.; Cai, C.; Wexler, J.; Viegas, F.; et~al. 2018.
\newblock Interpretability beyond feature attribution: Quantitative testing with concept activation vectors (tcav).
\newblock In \emph{Proceedings of the International Conference on Machine Learning}, 2668--2677.

\bibitem[{Koh et~al.(2020)Koh, Nguyen, Tang, Mussmann, Pierson, Kim, and Liang}]{koh2020concept}
Koh, P.~W.; Nguyen, T.; Tang, Y.~S.; Mussmann, S.; Pierson, E.; Kim, B.; and Liang, P. 2020.
\newblock Concept bottleneck models.
\newblock In \emph{International conference on machine learning}, 5338--5348. PMLR.

\bibitem[{K{\"u}sters et~al.(2020)K{\"u}sters, Schichtel, Ahmed, and Dengel}]{kusters2020conceptual}
K{\"u}sters, F.; Schichtel, P.; Ahmed, S.; and Dengel, A. 2020.
\newblock Conceptual explanations of neural network prediction for time series.
\newblock In \emph{2020 International joint conference on neural networks (IJCNN)}, 1--6. IEEE.

\bibitem[{Lam et~al.(2021)Lam, Chu, Torgonskiy, Pei, Zhang, and Wang}]{lam2021finding}
Lam, P. C.-H.; Chu, L.; Torgonskiy, M.; Pei, J.; Zhang, Y.; and Wang, L. 2021.
\newblock Finding representative interpretations on convolutional neural networks.
\newblock In \emph{Proceedings of the IEEE/CVF International Conference on Computer Vision}, 1345--1354.

\bibitem[{LeCun, Bengio, and Hinton(2015)}]{lecun2015deep}
LeCun, Y.; Bengio, Y.; and Hinton, G. 2015.
\newblock Deep learning.
\newblock \emph{nature}, 521(7553): 436--444.

\bibitem[{Montufar et~al.(2014)Montufar, Pascanu, Cho, and Bengio}]{montufar2014number}
Montufar, G.~F.; Pascanu, R.; Cho, K.; and Bengio, Y. 2014.
\newblock On the number of linear regions of deep neural networks.
\newblock \emph{Proceedings of the Advances in neural information processing systems}, 27.

\bibitem[{Nauta et~al.(2023{\natexlab{a}})Nauta, Schl{\"o}tterer, van Keulen, and Seifert}]{nauta2023pip}
Nauta, M.; Schl{\"o}tterer, J.; van Keulen, M.; and Seifert, C. 2023{\natexlab{a}}.
\newblock PIP-Net: Patch-Based Intuitive Prototypes for Interpretable Image Classification.
\newblock In \emph{Proceedings of the IEEE/CVF Conference on Computer Vision and Pattern Recognition}, 2744--2753.

\bibitem[{Nauta et~al.(2023{\natexlab{b}})Nauta, Trienes, Pathak, Nguyen, Peters, Schmitt, Schl{\"o}tterer, van Keulen, and Seifert}]{nauta2023anecdotal}
Nauta, M.; Trienes, J.; Pathak, S.; Nguyen, E.; Peters, M.; Schmitt, Y.; Schl{\"o}tterer, J.; van Keulen, M.; and Seifert, C. 2023{\natexlab{b}}.
\newblock From anecdotal evidence to quantitative evaluation methods: A systematic review on evaluating explainable ai.
\newblock \emph{ACM Computing Surveys}, 55(13s): 1--42.

\bibitem[{Oikarinen and Weng(2023)}]{oikarinen2022clip}
Oikarinen, T.; and Weng, T.-W. 2023.
\newblock CLIP-Dissect: Automatic Description of Neuron Representations in Deep Vision Networks.
\newblock arXiv:2204.10965.

\bibitem[{Sajjad, Durrani, and Dalvi(2022)}]{sajjad2022neuron}
Sajjad, H.; Durrani, N.; and Dalvi, F. 2022.
\newblock Neuron-level interpretation of deep nlp models: A survey.
\newblock \emph{Transactions of the Association for Computational Linguistics}, 10: 1285--1303.

\bibitem[{Samek et~al.(2019)Samek, Montavon, Vedaldi, Hansen, and M{\"u}ller}]{samek2019explainable}
Samek, W.; Montavon, G.; Vedaldi, A.; Hansen, L.~K.; and M{\"u}ller, K.-R. 2019.
\newblock \emph{Explainable AI: interpreting, explaining and visualizing deep learning}, volume 11700.
\newblock Springer Nature.

\bibitem[{Sandler et~al.(2018)Sandler, Howard, Zhu, Zhmoginov, and Chen}]{sandler2018mobilenetv2}
Sandler, M.; Howard, A.; Zhu, M.; Zhmoginov, A.; and Chen, L.-C. 2018.
\newblock Mobilenetv2: Inverted residuals and linear bottlenecks.
\newblock In \emph{Proceedings of the IEEE conference on computer vision and pattern recognition}, 4510--4520.

\bibitem[{Schrouff et~al.(2021)Schrouff, Baur, Hou, Mincu, Loreaux, Blanes, Wexler, Karthikesalingam, and Kim}]{schrouff2021best}
Schrouff, J.; Baur, S.; Hou, S.; Mincu, D.; Loreaux, E.; Blanes, R.; Wexler, J.; Karthikesalingam, A.; and Kim, B. 2021.
\newblock Best of both worlds: local and global explanations with human-understandable concepts.
\newblock \emph{arXiv preprint arXiv:2106.08641}.

\bibitem[{Selvaraju et~al.(2017)Selvaraju, Cogswell, Das, Vedantam, Parikh, and Batra}]{selvaraju2017grad}
Selvaraju, R.~R.; Cogswell, M.; Das, A.; Vedantam, R.; Parikh, D.; and Batra, D. 2017.
\newblock Grad-cam: Visual explanations from deep networks via gradient-based localization.
\newblock In \emph{Proceedings of the IEEE international conference on computer vision}, 618--626.

\bibitem[{Simonyan, Vedaldi, and Zisserman(2014)}]{simonyan2014visualising}
Simonyan, K.; Vedaldi, A.; and Zisserman, A. 2014.
\newblock Visualising image classification models and saliency maps.
\newblock \emph{Deep Inside Convolutional Networks}, 2.

\bibitem[{Simonyan and Zisserman(2014)}]{simonyan2014very}
Simonyan, K.; and Zisserman, A. 2014.
\newblock Very deep convolutional networks for large-scale image recognition.
\newblock \emph{arXiv preprint arXiv:1409.1556}.

\bibitem[{Sundararajan, Taly, and Yan(2017)}]{sundararajan2017axiomatic}
Sundararajan, M.; Taly, A.; and Yan, Q. 2017.
\newblock Axiomatic attribution for deep networks.
\newblock In \emph{International conference on machine learning}, 3319--3328. PMLR.

\bibitem[{Vinyals et~al.(2016)Vinyals, Blundell, Lillicrap, Wierstra et~al.}]{vinyals2016matching}
Vinyals, O.; Blundell, C.; Lillicrap, T.; Wierstra, D.; et~al. 2016.
\newblock Matching networks for one shot learning.
\newblock \emph{Advances in neural information processing systems}, 29.

\bibitem[{Zhao and Karypis(2001)}]{zhao2001criterion}
Zhao, Y.; and Karypis, G. 2001.
\newblock Criterion functions for document clustering: Experiments and analysis.

\end{thebibliography}

\clearpage

\onecolumn
\section*{\LARGE\centering\textbf{APPENDIX}}
\begin{center}
\line(1,0){\textwidth}
\end{center}

\section{Proofs for the Greedy Algorithm}

\begin{lemma}
    Let $z_1$, $z_2$ be $m$-categorical random variables, i.e., $z_1,z_2\in\{1,\cdots,m\}$. Given a state $p\in\{1,\cdots,m\}$, consider the following minimization problem
    \begin{equation}
    \label{eq:obj}
        \underset{p\in\{1,\cdots,m\}}{\mathrm{argmin}}
        \;\mathbb{E}_{z_1}[\mathbf{1}_{[z_1\neq p]}]-\mathbb{E}_{z_2}[\mathbf{1}_{[z_2\neq p]}]
    \end{equation}
    where $\mathbf{1}_{[\cdot]}$ denotes the indicator function.
    Then, the optimal solution is
    \begin{equation}
    \label{eq:optsol}
        p^* = \underset{p\in\{1,\cdots,m\}}{\mathrm{argmax}}
        \;\mathbb{P}(z_1=p)-\mathbb{P}(z_2=p).
    \end{equation}
\end{lemma}
\begin{proof}
For a fixed $p\in \{1,\cdots,m\}$, 
\begin{equation*}
    \begin{aligned}
        &\quad \mathbb{E}_{z_1}[\mathbf{1}_{[z_1\neq p]}]-\mathbb{E}_{z_2}[\mathbf{1}_{[z_2\neq p]}] \\
        &= \sum_{p'\in\{1,\cdots,m\}}
        \mathbb{P}(z_1=p')\mathbf{1}_{[p'\neq p]} - \sum_{p'\in\{1,\cdots,m\}}\mathbb{P}(z_2=p')\mathbf{1}_{[p'\neq p]}
        \\
        &= \sum_{p'\in\{1,\cdots,m\}\setminus \{p\}} \Big( 
        \mathbb{P}(z_1=p') - \mathbb{P}(z_2=p') \Big)
        \\
        &= 2 - \Big( \mathbb{P}(z_1=p) - \mathbb{P}(z_2=p) \Big)
    \end{aligned}
\end{equation*}
Therefore, the minimization problem in Equation (\ref{eq:obj}) has the optimal value at $p=p^*$.
\end{proof}

From Lemma 1, we prove Theorem 1 of the main paper.
\begin{proof}[proof of Theorem 1]
Let $\mathbf{x}$ and $\mathbf{y}$ be the random variables corresponding to the positive set and the negative set, respectively.
\begin{equation*}
    \begin{aligned}
        &\underset{{\underset{\scriptstyle N^*\subset N}{\mathbf{c}_{\mathrm{p}}\in {\{0,1\}}^t}}}{\mathrm{argmin}}
         \;\mathbb{E}_{\mathbf{x}}\Big[d_H(\mathbf{c}^{N^*}(\mathbf{x}),\mathbf{c}_{\mathrm{p}})\Big] - 
\mathbb{E}_{\mathbf{y}}\Big[
d_H(\mathbf{c}^{N^*}(\mathbf{y}),\mathbf{c}_{\mathrm{p}})\Big]
    \\ =&\underset{{\underset{\scriptstyle N^*\subset N}{\mathbf{c}_{\mathrm{p}}\in {\{0,1\}}^t}}}{\mathrm{argmin}} \;
    \mathbb{E}_{\mathbf{x}} 
    \Big[ \sum_{i\in N^*} \mathbf{1}_{[\mathbf{c}^N_i(\mathbf{x})\neq\mathbf{c}_{\mathrm{p},i}]}\Big] - 
    \mathbb{E}_{\mathbf{y}}\Big[\sum_{i\in N^*}    \mathbf{1}_{[\mathbf{c}^N_i(\mathbf{y})\neq\mathbf{c}_{\mathrm{p},i}]}\Big]
    \\ =&\underset{{\underset{\scriptstyle N^*\subset N}{\mathbf{c}_{\mathrm{p}}\in {\{0,1\}}^t}}}{\mathrm{argmin}} \; \sum_{i\in N^*}\Big\{
    \mathbb{E}_{\mathbf{x}} 
    [  \mathbf{1}_{[\mathbf{c}^N_i(\mathbf{x})\neq\mathbf{c}_{\mathrm{p},i}]}] - 
    \mathbb{E}_{\mathbf{y}}[    \mathbf{1}_{[\mathbf{c}^N_i(\mathbf{y})\neq\mathbf{c}_{\mathrm{p},i}]}] \Big\}
    \\ =&\underset{\scriptstyle N^*\subset N}{\mathrm{argmax}} \; \sum_{i\in N^*}\Big\{
    \max_{\mathbf{c}_{\mathrm{p},i}\in \{1,\cdots,m\}}
    \mathbb{P}(\mathbf{c}_i(\mathbf{x})=\mathbf{c}_{\mathrm{p},i})-
    \mathbb{P}(\mathbf{c}_i(\mathbf{y})=\mathbf{c}_{\mathrm{p},i})
    \Big\}
    \qquad\qquad(\because \textrm{ Lemma 1})
    \end{aligned}
\end{equation*}
It means that $\mathbf{c}_{\mathrm{p},i}$ is set to the state that maximizes the frequency difference between two distributions. Note that there is a constraint for the number of neurons to select in the objective of the original problem ($|N|=t$). Therefore, the optimal solution can be obtained by iteratively selecting a neuron $i$ that maximizes
\begin{equation}
\label{eq:expect}
    \max_{\mathbf{c}_{\mathrm{p},i}\in \{1,\cdots,m\}}
    \mathbb{P}(\mathbf{c}^N_i(\mathbf{x})=\mathbf{c}_{\mathrm{p},i})-
    \mathbb{P}(\mathbf{c}^N_i(\mathbf{y})=\mathbf{c}_{\mathrm{p},i}).
\end{equation}

Consider the ReLU activation function that has 2 activation states ($\{0,1\}$), the positive set $S$ and the negative set $S_{\textrm{neg}}$ from the training data. Let $\bar{\textbf{c}}$, $\bar{\textbf{c}}_{\mathrm{neg}}$ be the frequency of neuron activations for instances in $S$ and $S_\mathrm{neg}$, respectively.
\begin{equation*}
\begin{aligned}
   &\bar{\textbf{c}}=\frac{1}{|S|} \sum_{\mathbf{x}\in S}
        \mathbf{c}^N(\mathbf{x}) , \quad \bar{\textbf{c}}_{\mathrm{neg}}=\frac{1}{|S_{\mathrm{neg}}|} \sum_{\mathbf{y}\in S_{\mathrm{neg}}}
        \mathbf{c}^N(\mathbf{y}).
\end{aligned}
\end{equation*}

Then, Equation (\ref{eq:expect}) can be represented as
\begin{equation*}
\begin{aligned}
    \mathbf{c}_{\mathrm{p},i}=\underset{p\in\{0,1\}}{\mathrm{argmax}}\;
    \frac{1}{|S|}\sum_{\mathbf{x}\in S}\mathbf{1}_{[\mathbf{c}^N_i(\mathbf{x})=p]}
    - \frac{1}{|S_\mathrm{neg}|}\sum_{\mathbf{y}\in S_{\mathrm{neg}}}\mathbf{1}_{[\mathbf{c}^N_i(\mathbf{y})=p]}\\
\end{aligned}
\end{equation*}
Consider two cases: 1) If $\bar{\mathbf{c}}_i < \bar{\mathbf{c}}_{\mathrm{neg},i}$, then $\mathbf{c}_{\mathrm{p},i}=0$, 2) If $\bar{\mathbf{c}}_i \geq \bar{\mathbf{c}}_{\mathrm{neg},i}$, then $\mathbf{c}_{\mathrm{p},i}=1$.
\begin{equation*}
\textrm{the objective}=
    \begin{cases}
(1-\mathbf{\bar{c}}_i) - (1-\bar{\mathbf{c}}_{\mathrm{neg},i})
    = -\mathbf{\bar{c}}_i + \bar{\mathbf{c}}_{\mathrm{neg},i}
    = |\mathbf{\bar{c}}_i - \bar{\mathbf{c}}_{\mathrm{neg},i}| \quad& \mathrm{\;if}\; \bar{\mathbf{c}}_i < \bar{\mathbf{c}}_{\mathrm{neg},i} \\
\mathbf{\bar{c}}_i - \bar{\mathbf{c}}_{\mathrm{neg},i}
    = |\mathbf{\bar{c}}_i - \bar{\mathbf{c}}_{\mathrm{neg},i}| 
\quad& \mathrm{\;if}\; \bar{\mathbf{c}}_i \geq \bar{\mathbf{c}}_{\mathrm{neg},i}.
    \end{cases}
\end{equation*}
Therefore, the optimal solution $N^*$ is a set of $t$ neurons that consists of the top $t$ values of $|\mathbf{\bar{c}}_i - \bar{\mathbf{c}}_{\mathrm{neg},i}|$.
\end{proof}

\section{Experimental Details}

This section describes the experimental details including datasets, models, used algorithms, and hyperparameters.

\subsection{Datasets}
We use 6 datasets to verify the performance of our framework. We organize the detailed information of each dataset below.

{\renewcommand{\arraystretch}{1.0}
\begin{table}[h]
\centering
\label{dataset}
\begin{tabularx}{\textwidth}{c|cccccc}
\noalign{\smallskip}\noalign{\smallskip}\hline\hline
 & Mini-ImageNet & Flowers Recognition & CIFAR-10 & Oxford-Pet & Broden & ImageNet-X \\
\hline
Resolution size & 84*84 & 224*224 & 28*28 & 84*84 & 224*224 & 224*224 \\
Number of labels & 64 & 5 & 10 & 2 (37) & - & 64 (1000) \\
Number of samples & 38400 & 4242 & 50000 & 7349 & 63305 & 3700 (50000) \\
\hline
\hline
\end{tabularx}
\end{table}
}

The Mini-ImageNet~\citep{vinyals2016matching} and the Flowers Recognition\footnote{\url{https://www.kaggle.com/datasets/alxmamaev/flowers-recognition}} dataset are typical datasets for image classification tasks and we use them without preprocessing. The original Oxford-Pet\footnote{\url{https://www.robots.ox.ac.uk/~vgg/data/pets/}} dataset consists of 37 classes (breeds of dog, cat), but we combine them into two classes (dog and cat) for conducting subclass detection experiments. In the Broden dataset~\cite{bau2017network}, there are pixel-wise segmentation labels such as color, object, scene, and texture instead of per-image class labels. We use this dataset to measure the alignment between captured concepts with our methods and a set of pre-defined concepts. ImageNet-X~\citep{idrissi2022imagenet} is a set of human annotations pinpointing failure types for the ImageNet dataset. ImageNet-X labels distinguishing object factors such as pose, size, color, lighting, occlusions, co-occurrences, etc. The original ImageNet-X has 50000 samples with 1000 labels but according to the purpose of our experiment, we extract images corresponding to the labels of the Mini-ImageNet dataset. Then, there are 3700 images with 64 labels. 

\subsection{Models}
We use CNN6, VGG19, ResNet50, and MobileNetV2 models to verify our algorithm. These models consist of convolution layers, linear layers, residual connections, average pooling, max pooling, and piecewise linear activation functions. These operations make trained neural networks locally linear in the input space and the intermediate feature space made. In this setting, having identical activation states induces the same decision mechanism in the model, which means that the model extracts the same information from data.



In the case of VGG19, we fine-tune the pre-trained model for each dataset, adjusting the dimensions of the linear layers according to the data. We use linear layers with either 512 or 1024 dimensions. ResNet follows the structure of the pre-trained (trained on ImageNet) model, similar to the VGG19 setting, we further add linear layers to the vanilla ResNet50 structure which has a single linear layer. As for MobileNetV2, we utilize the pre-trained (with ImageNet) model. The Convolutional Neural Network (CNN) that we use comprises 6 convolutional layers and 2 linear layers. Each of the models achieved a training accuracy between 0.9 and 1, as well as a test accuracy ranging from 0.7 to 0.9. Both the features and the linear classifiers are well-trained in the models. Therefore, these models can be interpreted through our method.


For all experiments except the layer dynamics experiment, we set the target layer as the penultimate convolution block emphasizing semantically evident, non-class-specific features: the 4th layer of CNNs, the 12th layer of VGG19 models, the 3rd residual block of ResNet50, and the 7th layer of MobileNetV2.
 
\subsection{Relaxed Decision Region}
Our algorithm forms an interpretable region, called Relaxed Decision Region (RDR), which contains semantically similar instances that share concepts learned in the target layers. To constitute an appropriate RDR, we select the principal neurons that represent the concept. In this process, we set the positive (concept) set and the negative (concept) set to evaluate neurons.

\begin{figure}[ht!]
\centering
\includegraphics[width=\columnwidth]{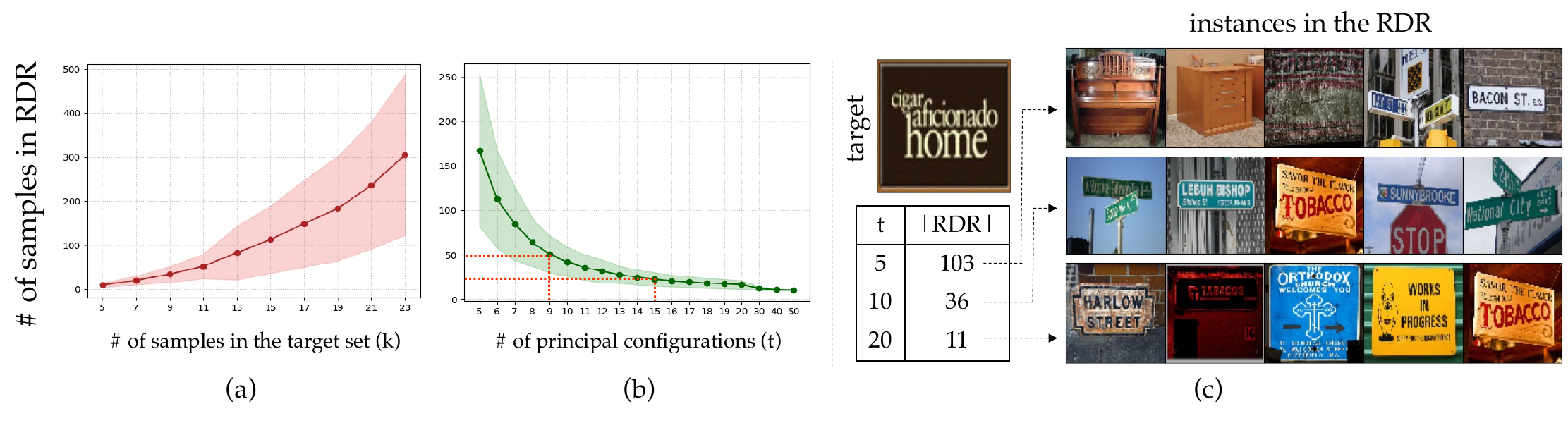}
\caption{Design choices made in the RDR. We set $t$ = 10 in (a) and set $k$ = 10 in (b) and (c).}
\label{fig:ablation}
\end{figure}

\textbf{Design Choices.} \; As we also described in the main paper, there are 2 controllable parameters, $t$ and $k$. $k$ is a parameter that determines the size of the positive set. As the $k$ value increases, the proposed algorithm should consider the configurations shared by more instances. Therefore, concepts with more general characteristics will be extracted, which leads to an increase in the number of instances in the RDR. Figure~\ref{fig:ablation}-(a) shows the relation between $k$ and the tightness of the RDR. As $k$ increases, the variance of the properties in the positive set increases. However, it is not very sensitive when $k$ is set under 10. We recommend that the user set $k \in [5,10]$ to form the adequate Relaxed Decision Region. $t$ is the number of neurons in the principal configuration. If $t$ increases, the RDR becomes tighter as shown in Figure~\ref{fig:ablation}-(b-c). Since we empirically observe that at least more than 9 principal neurons are usually needed to represent a coherent and distinct concept regardless of the target sample, we recommend a $t \in [9,15]$. When set this way, the number of instances included within the RDR is typically 25 to 50. The analyses in Figure~\ref{fig:ablation} are conducted in the penultimate convolution block of VGG19. The required $t$ and $k$ values may vary slightly when verifying the concepts of different layers because of the sparsity difference within the feature space.

\textbf{Computational Cost}
The major computational cost arises from the essential forward pass for collecting data configuration, taking approximately 15 seconds for VGG19 on Mini-ImageNet with a single GPU (Quadro RTX 6000). It is indispensable for any group-level interpretation method using feature maps, such as ACE. The subsequent steps, involving Configuration distance computation and RDR execution,  take less than 2 seconds (easy to test RDR multiple times). The overall time complexity of the process is $O(N)$. Even the computation of saliency maps, for providing a group-level explanation, also requires $O(N)$.


\subsection{Baseline Methods}

\textbf{Representative Interpretation.}\; Representative interpretation (RI)~\citep{lam2021finding}  provides linear decision boundaries that form co-clusters in such a way that the number of instances with the same class in each cluster is maximized. The algorithm allows for control over the number of linear boundaries to be computed as a hyperparameter. Here, we expressed by computing, not selecting because RI finds the boundary based on the point where the logit changes, which may not correspond exactly to the actual internal boundary. This implies that computing these linear boundaries requires heavy computational costs. In experiments, we permit a sufficiently large number of candidates for linear boundaries to get appropriate results (we set to 300). However, as shown in the experimental results, they are not good at finding concepts like subclasses, because they are optimized in the direction of increasing the number of samples with the same class.



\textbf{CAVs.} \; \citet{kim2018interpretability} represent the learned concept with the fixed direction of the DNN's feature space, denoted as a concept activation vector (CAV). They find a vector in which the difference between the concept positive set and the concept negative set is maximized. The approach is similar to ours when they find a vector in the feature space rather than directly find the subset of neurons and they require a pre-defined concept positive set. 

\textbf{ACE.} \;
Borrowing the idea of CAVs, \citet{ghorbani2019towards} propose an automatic method, called Automated Concept-based Explanation (ACE), to overcome the dependency on the existing pre-defined concept set. They apply the segmentation process to make segmentation patches as a pre-defined concept set and calculate the TCAV score with a cluster of these patched images to find the parts that are important for classification. Similar to our approach, ACE also finds the concept set in an unsupervised way, while it differs from ours in that it uses patched images and focuses on class-specific interpretations.

\textbf{CAR.} \; Contrary to CAVs that represent concepts as vectors, \citet{crabbé2022concept} identify non-linear concept regions, namely Concept Activation Regions (CARs) in the DNN's feature space. They employ a kernel Support Vector Machine (kernel-SVM) with a Radial Basis Function (RBF) kernel. A hyperparameter used in their SVM model is the kernel width, which plays a pivotal role in controlling the complexity of the learned decision boundaries. They optimize the kernel width using the Optuna hyperparameter optimization framework with 1000 trials. In our experiments, we adhere to the settings employed by CARs, maintaining consistency with their original settings and thereby enabling a fair comparison of results. Note that CARs also need a pre-defined concept positive set. 

\textbf{K-Nearest Neighborhoods.}\; 
For fair comparisons, we explore the nearest instances for each distance metric, ensuring an equal number of instances within a Relaxed Decision Region.

\textbf{Grad-CAM.} \;
Grad-CAM is a visualization technique that shows which parts a CNN model focuses on by using feature map information~\cite{selvaraju2017grad}. In the convolutional layer, it evaluates the importance of each channel by computing its gradient to the final logit. Even though Grad-CAM is a popular visualization method, it does not guarantee consistent results in the low or middle layers of CNNs.


\textbf{IG.} \; Integrated Gradients (IG)~\cite{sundararajan2017axiomatic} is one of the input attribution methods that compute the attribution of input variables to the final logit. It computes the cumulative sum of the gradients from the base point to the original input. In our experiments, we compute IG for the feature map of the intermediate feature space by setting a zero vector as a base point. While IG usually captures the object in the image, it may not provide consistent heatmaps to the group of similar images.

\begin{figure}[ht!]
\begin{center}
\centerline{\includegraphics[width=0.85\columnwidth]{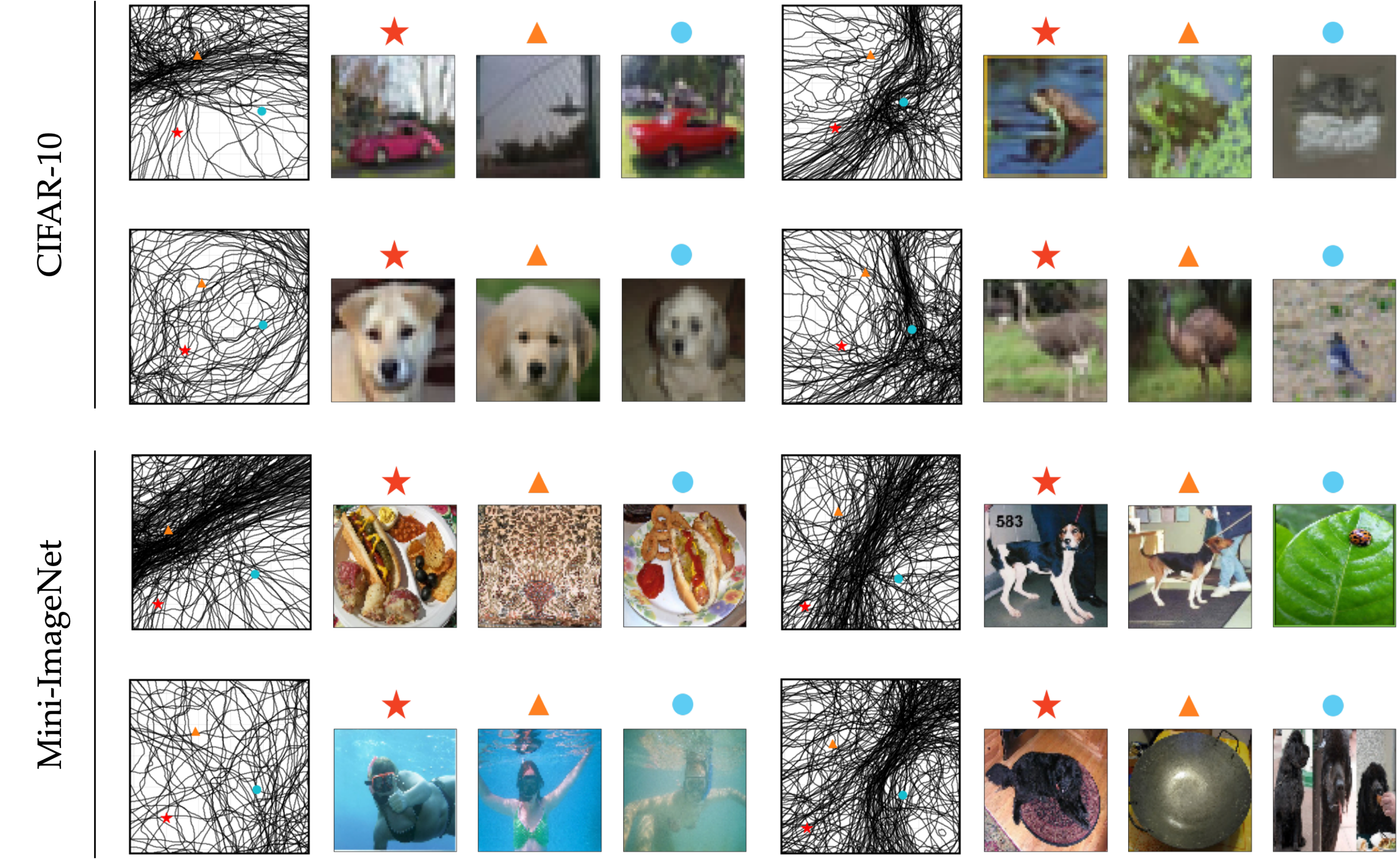}}
\caption{Visualization of the Feature space with Decision Boundaries on the CIFAR-10 dataset, and the Mini-ImageNet dataset.}
\label{fig:append-db-vis}
\end{center}
\end{figure}

\begin{figure}[ht!]
\centering
\includegraphics[width=0.85\columnwidth]{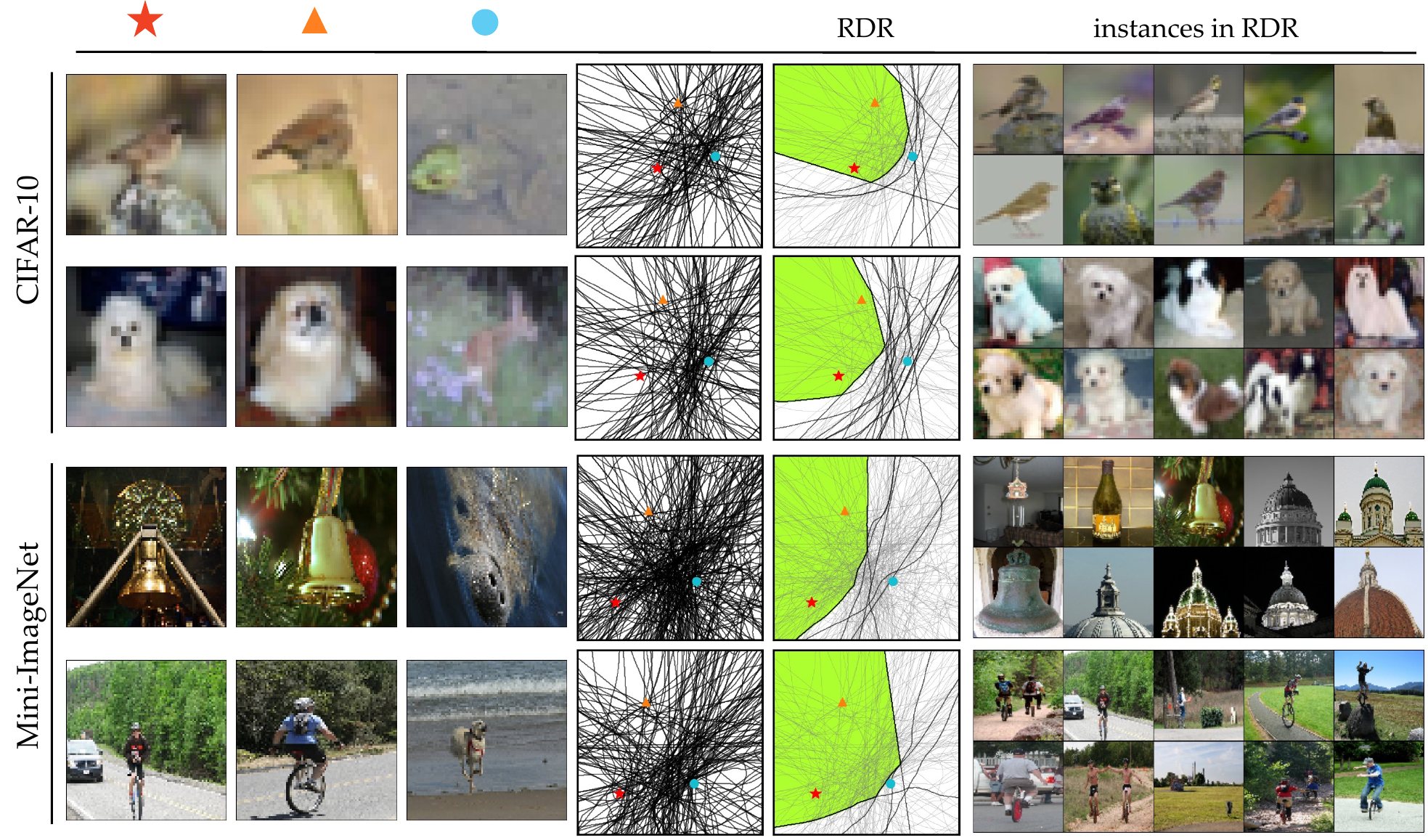}
\caption{Visualization of the Feature space with the Decision Boundaries and the Principal Configuration.}
\label{fig:append-pc-vis}
\end{figure}

\section{Decision Boundaries in the Feature Space}
In Figure~\ref{fig:append-db-vis}, we present more examples that visualize internal decision boundaries in the feature space. For three instances, we compute the plane that passes through them and draw decision boundaries on the 2-d plot. The selected layer is the penultimate layer. Each line on the plane represents an internal decision boundary corresponding to each neuron in the higher layers, i.e., two sides separated by a line have different states with respect to the neuron activation. We randomly select neurons for visualization since there are too many neurons to effectively display each boundary. 



\begin{figure}[ht!]
\centering
\includegraphics[width=0.95\columnwidth]{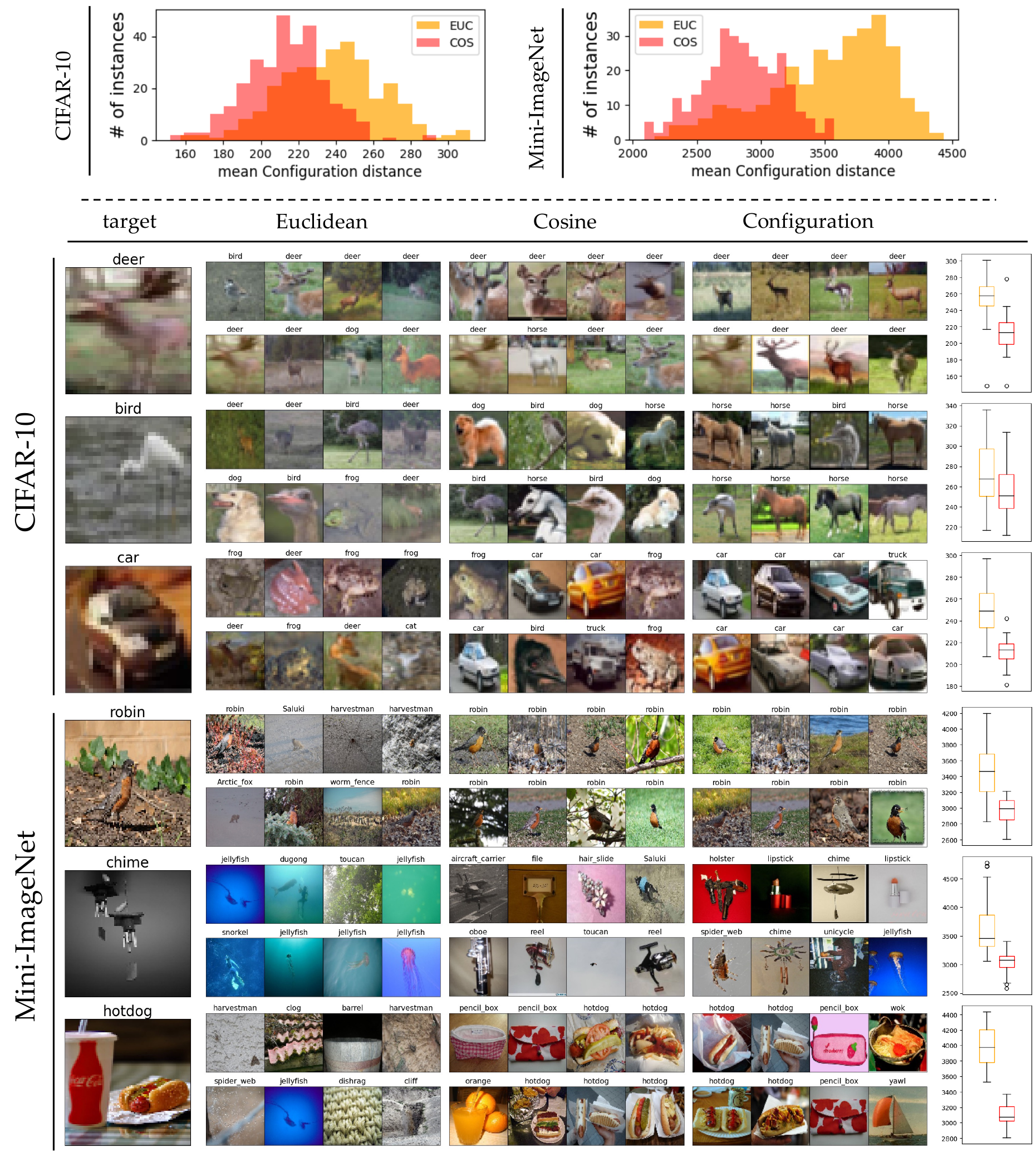}
\caption{Relationship of metrics in the feature space. }
\label{fig:append-euc-cos}
\end{figure}

As shown in Figure~\ref{fig:append-db-vis}, we can easily identify that internal decision boundaries are highly related to the semantics of instances. Even though instances have the same class, the semantic gap appears along with decision boundaries. When three instances share similar concepts, decision boundaries do not obviously separate the local feature space. The last case on the Mini-ImageNet has the dog image (red star) is misclassified to the wok. By observing decision boundaries, we can guess why the dog image is misclassified to wok.

\textbf{Relation to the Cosine Similarity.}\; The success of the Cosine similarity in the convolutional structures may highly attribute to estimating the difference in decision regions. Figure~\ref{fig:append-euc-cos} shows additional empirical evidence for this insight. For a random sample, we check the Configuration distance of 50-nearest neighbors based on the Euclidean distance and the Cosine similarity.

The above histograms show the average Configuration distance of neighbors across 500 trials. The Cosine similarity serves as a comparable measure to assess the semantic difference. The qualitative sample analysis is also depicted in Figure~\ref{fig:append-euc-cos}. The rightmost boxplots show that the mean and variance values are much smaller when using the Cosine similarity (represented in red) as compared to the Euclidean distance (represented in orange).

In Figure~\ref{fig:append-pc-vis}, we visualize the RDRs and encompassed instances. For each case, the first image (red star) is a target image to construct the RDR. Even though the third image (blue circle) is far from the first image than the second image (orange triangle), the third one has more dissimilar semantics. Also, there are numerous internal decision boundaries between them. The results support the necessity of the Configuration distance in the feature space of DNN. 

\section{Qualitative Comparison}

\begin{figure*}[hb!]
\centering
\includegraphics[width=0.95\columnwidth]{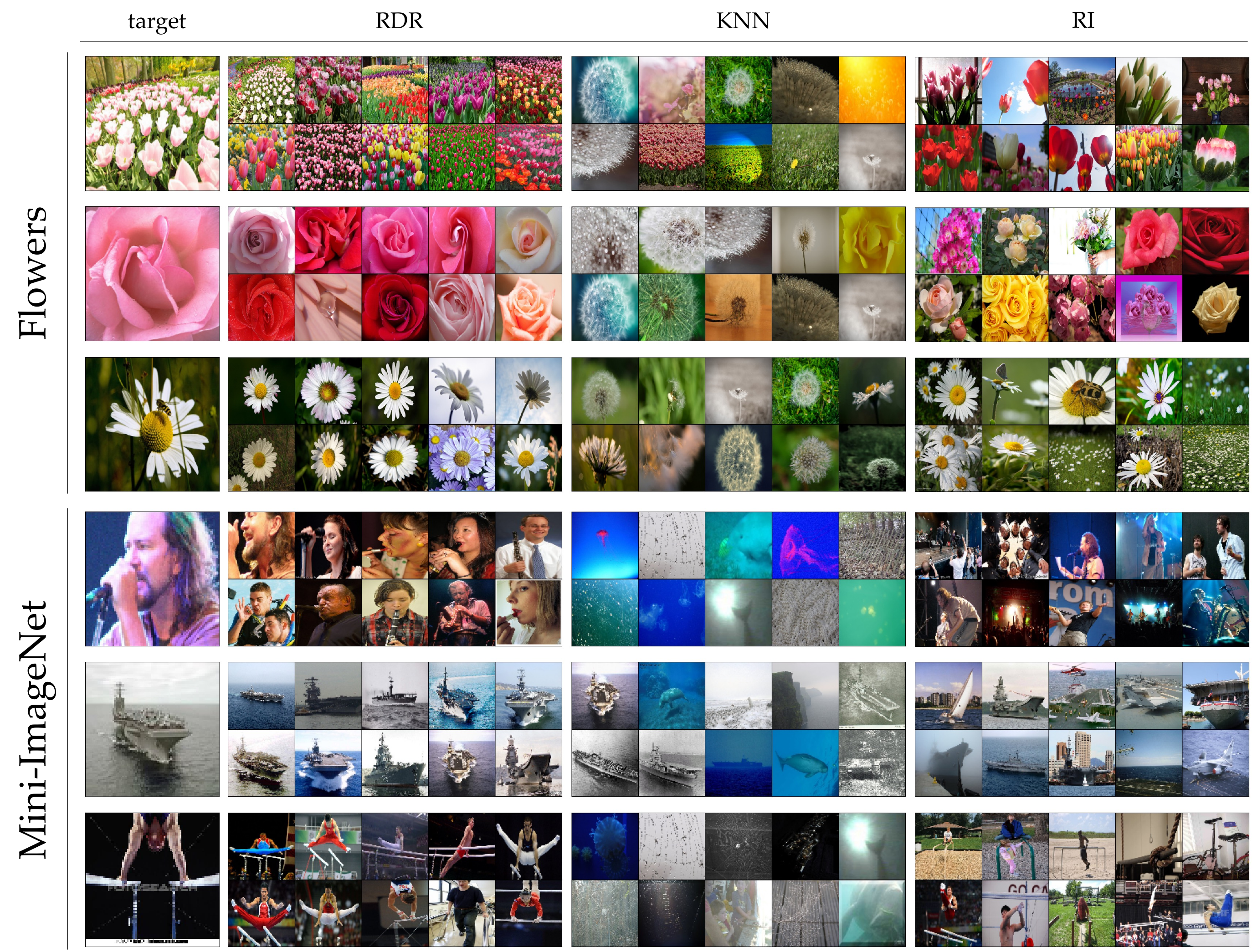}
\caption{Comparison with existing methods. We sampled 6 target instances and construct a concept-sharing group in 3 ways (K-Nearest Neighborhood method, Representative Interpretation method and ours). 10 samples in each concept-sharing group are randomly chosen for a fair comparison.} 
\label{fig:rdr_qual}
\end{figure*}

\begin{figure}[ht!] 
\centering
\centerline{\includegraphics[width=0.85\columnwidth]{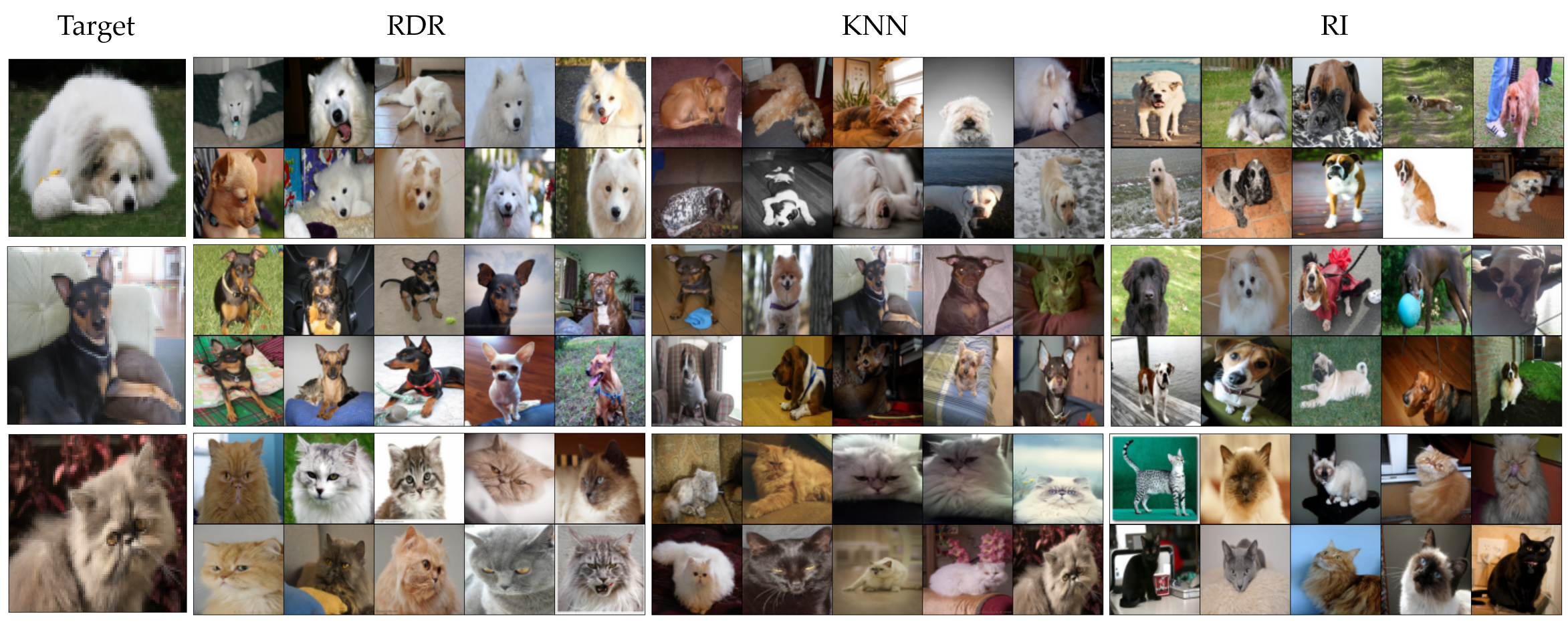}}
\caption{Relaxed Decision Region with Oxford-Pet dataset. The subclasses (breeds) are not given during the training procedure. } 
\label{fig:catdog_qual}
\end{figure}

To supplement the qualitative results in the main paper, we present qualitative comparisons in Mini-ImageNet, Flowers, and Oxford Pet datasets. Each dataset has 3 target instances and corresponding grouped instances that share learned concepts by each method. We compare the captured concept group of RDR, K-Nearest Neighborhood with Euclidean distance, and RI algorithm. Note that we do not qualitatively compare with CAV and CAR, because their vanilla settings require pre-defined concept set. In Figure~\ref{fig:rdr_qual} and Figure~\ref{fig:catdog_qual}, we empirically verify that the RDRs effectively capture the shared learned concepts the most. 

\section{Human Evaluation}
\begin{figure}[ht!] 
\centering
\centerline{\includegraphics[width=0.9\columnwidth]{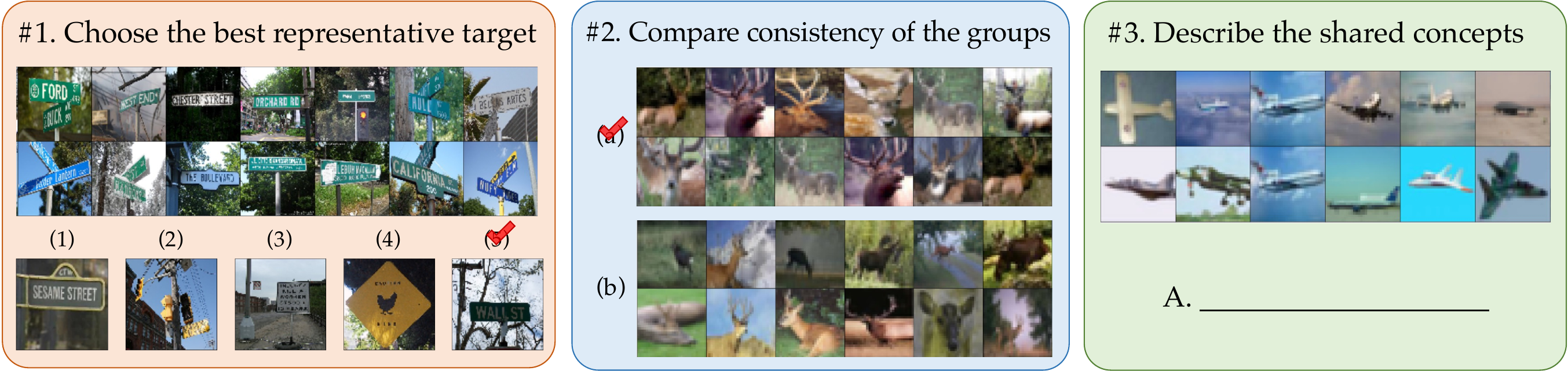}}
\caption{Questions on Human Evaluation.} 
\label{fig:human-eval-appx}
\end{figure}

We conduct a human-grounded evaluation to assess the consistency of captured concepts within groups, aligning with the Purity criterion discussed by~\citet{zhao2001criterion} and \citet{nauta2023anecdotal}. The survey was conducted with 22 participants and comprised three types of questions. Figure~\ref{fig:human-eval-appx} displays the example of each question type. The first question asked to identify the target instances when the randomly selected instances within RDRs are given. Remarkably, $81.8\%$ of participants easily identified which sample belongs to the given RDR group out of five samples with the same class. The second question compared the consistency of RDRs and the set grouped by other methods. $85.2\%$ of participants agreed that samples within RDR exhibit much clearer semantics compared to the other grouping methods. The last question required participants to describe the shared concepts among samples within RDR groups. In the case of the example in Figure~\ref{fig:human-eval-appx}, all participants recognized the airplane, and $59\%$ of them also mentioned the concept of sky background.

\section{Applications of Relaxed Decision Region}

\subsection{Application to the Various Intermediate Layers}
\label{sec:Layers}

\begin{figure}[hb!] 
\centering
\centerline{\includegraphics[width=0.75\columnwidth]{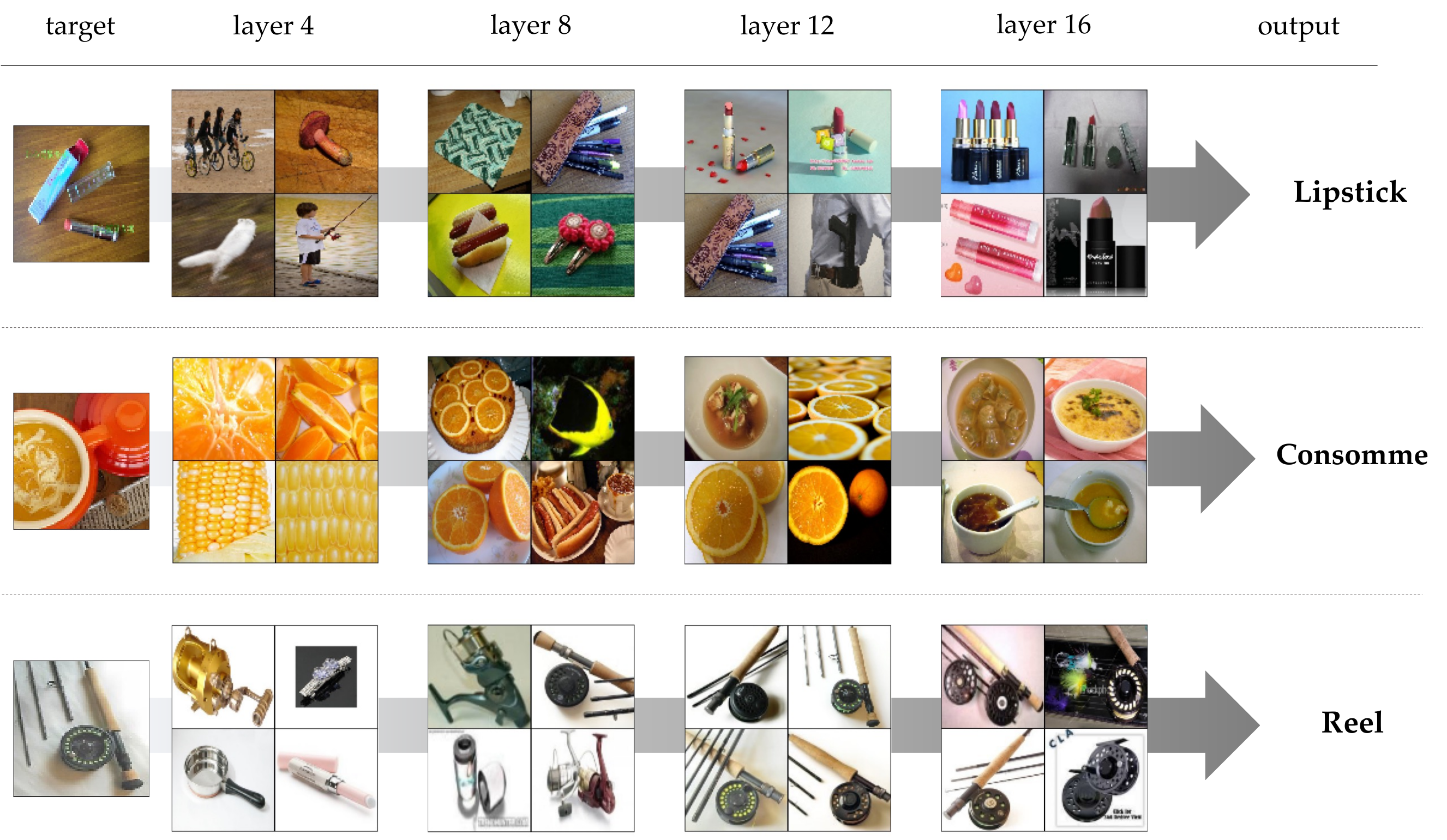}}
\caption{Layer-wise difference of Relaxed Decision Regions in VGG19: Different concepts are detected in each layer.} 
\label{fig:layers-appx}
\end{figure}

\begin{figure}[hb!] 
\centering
\centerline{\includegraphics[width=0.75\columnwidth]{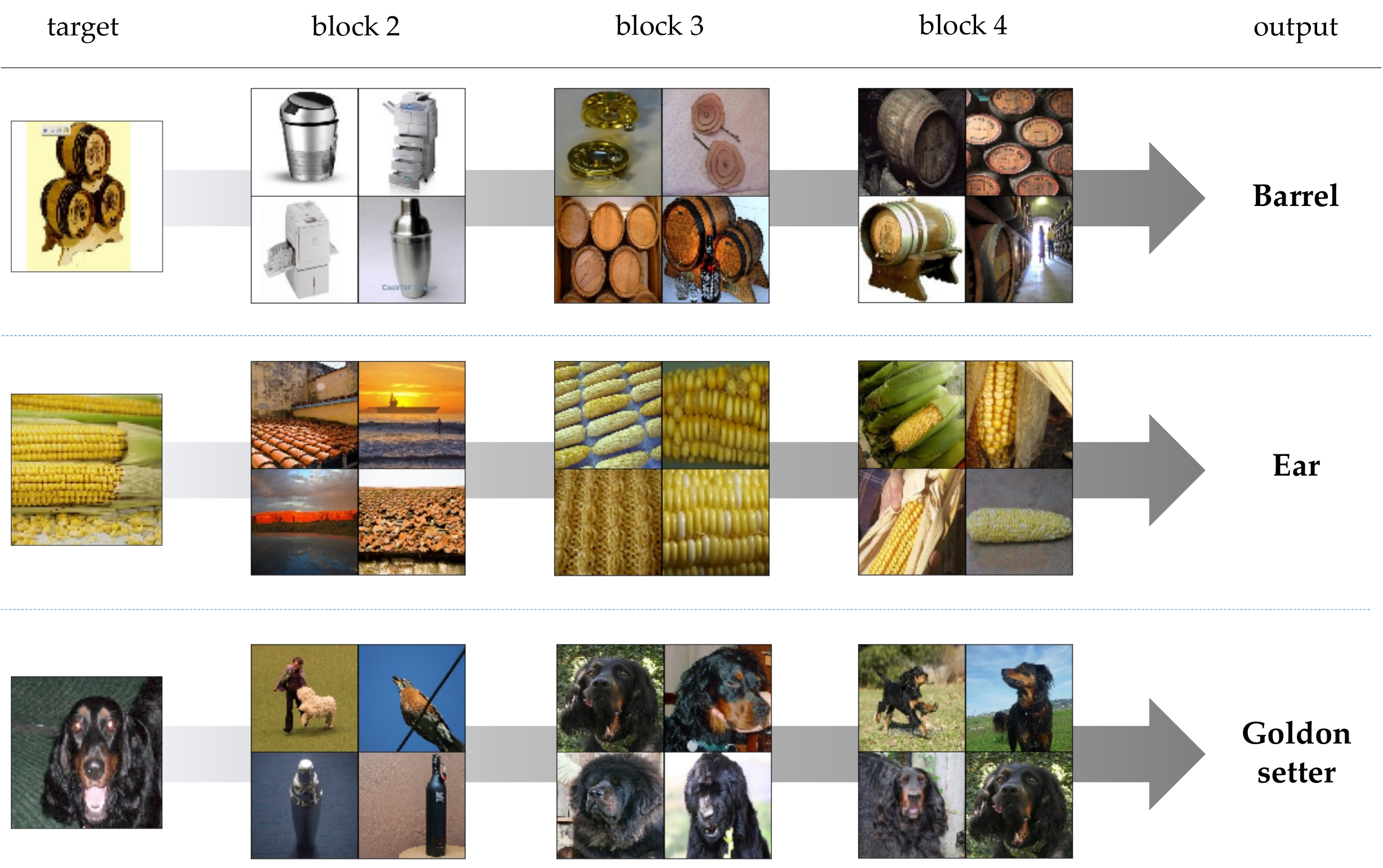}}
\caption{Layer-wise difference of Relaxed Decision Regions in ResNet50: Different concepts are detected in each layer.} 
\label{fig:layers-resnet}
\end{figure}

In this section, we show that the proposed algorithm can be applied to any intermediate layer. Since the complexity of the feature space is different according to the target layer, RDR can be differently constructed according to the layer even with the same target instance. It has been widely studied that the model learns different concepts across the layers~\cite{bau2017network}. We conduct the experiment with the VGG19 model. For the lipstick image in Figure~\ref{fig:layers-appx}, on the 4th layer, it seems that the long-shaped objects have been captured through our algorithm, while the captured learned concepts change into a more class-dependent direction as the layer goes deeper. In the last 16th layer, finally, the images in RDR appear to be dominant with class information. In Figure~\ref{fig:layers-resnet}, a similar phenomenon is observed in the ResNet50 model. 

\subsection{Application for debugging the model}
\label{misclass}



\begin{figure}[ht!] 
\centering
\centerline{\includegraphics[width=0.6\columnwidth]{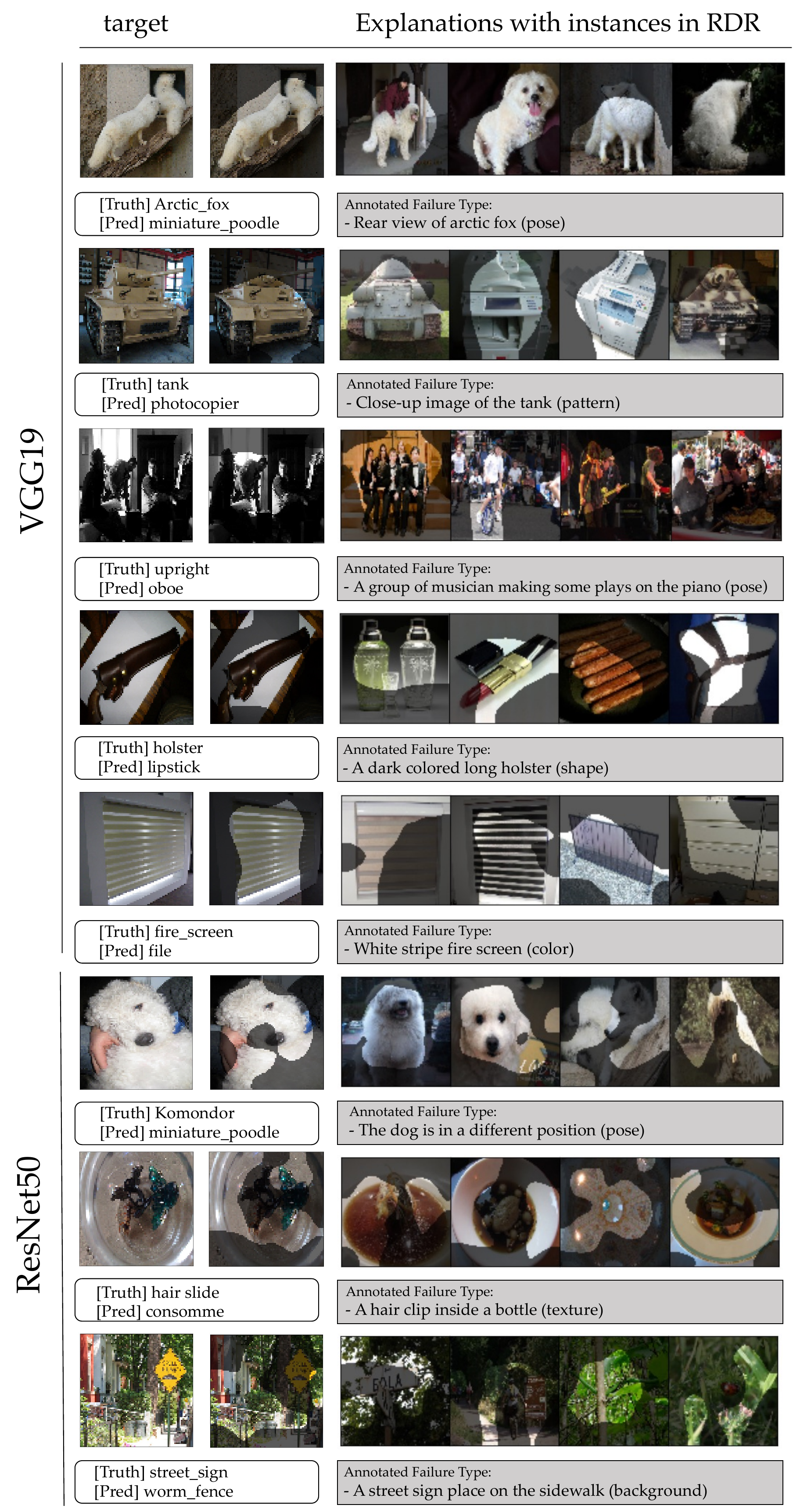}}
\caption{Relaxed Decision Regions with the misclassified target instances on ImageNet-X.} 
\label{fig:error_rdr}
\end{figure}

\begin{figure}[ht!] 
\centering
\centerline{\includegraphics[width=\columnwidth]{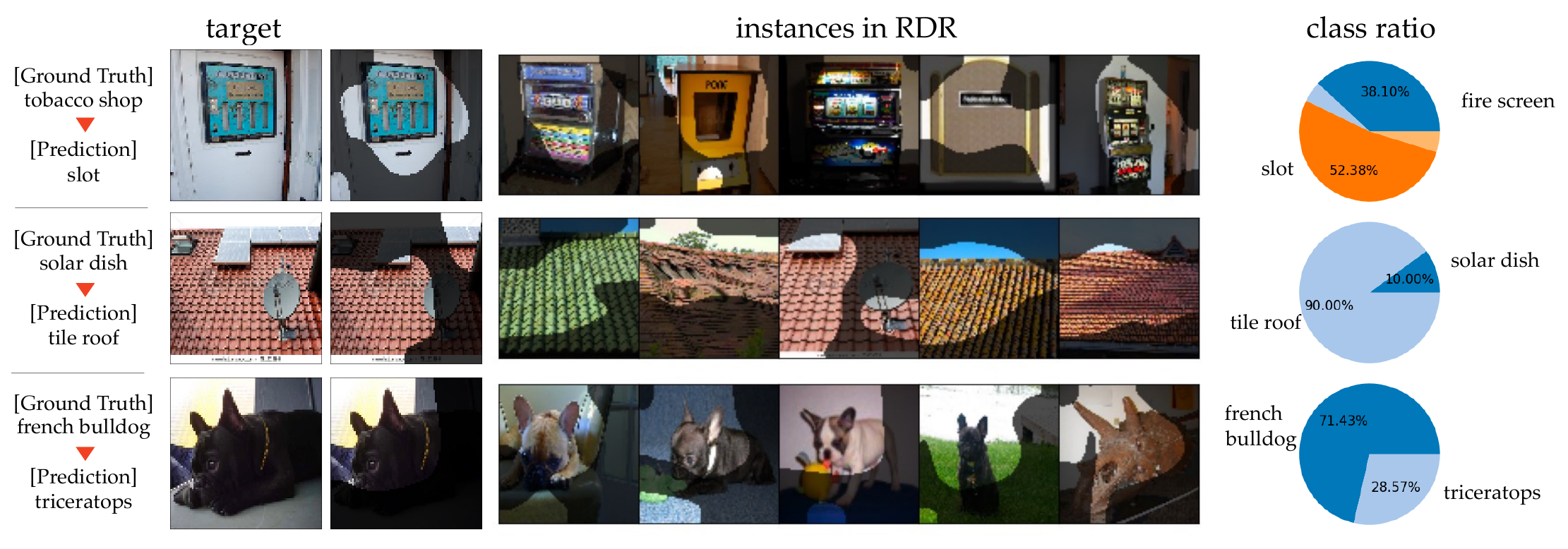}}
\caption{Misclassification detection with RDR} 
\label{fig:misclass-appx}
\end{figure}

\begin{figure}[ht!] 
\centering
\centerline{\includegraphics[width=0.9\columnwidth]{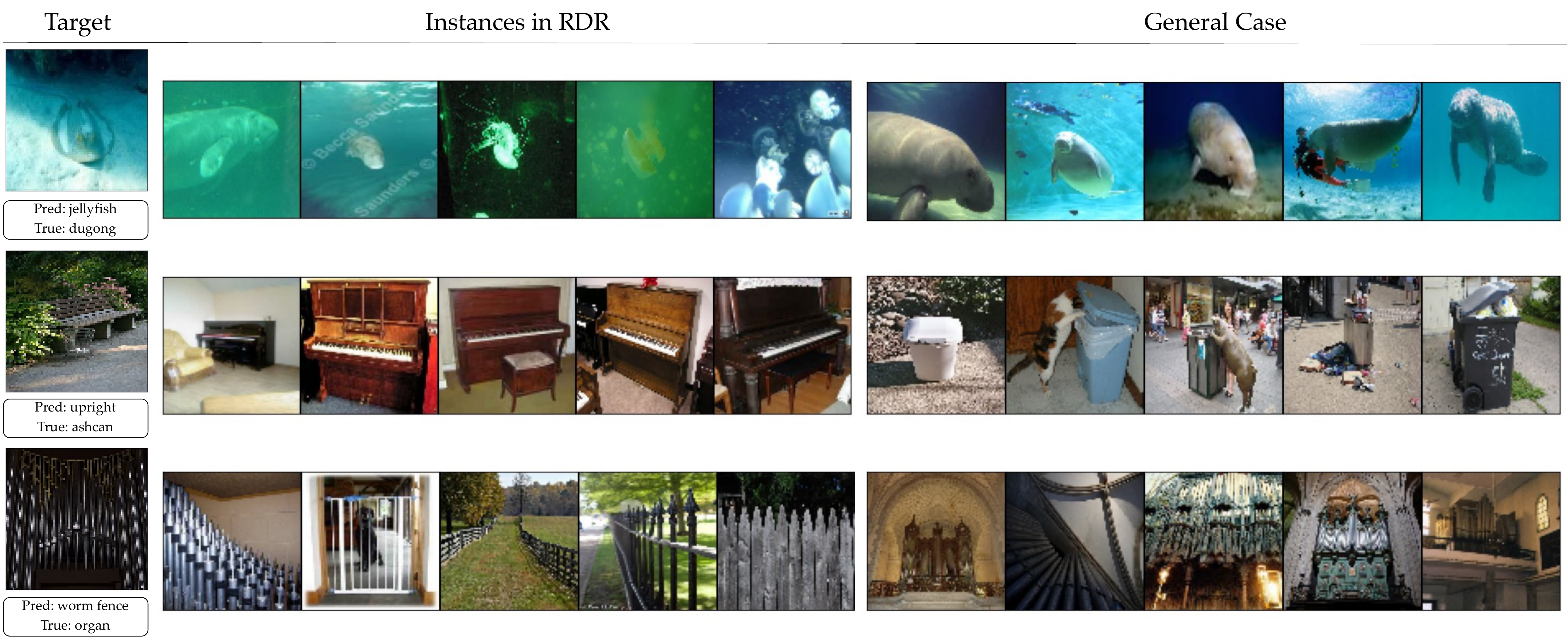}}
\caption{Relaxed Decision Regions with the misclassified target instances on ResNet50.} 
\label{fig:error_rdr_resnet}
\end{figure}

By manipulating the RDR, we can also detect the reason for the misclassification cases. If the instances with the true label of the misclassified target instance are taken as the reference set, we observe which concept causes the model to misclassify the target instance. As misclassification analysis in the main paper, we provide additional cases, including ResNet50, in Figure~\ref{fig:error_rdr} and Figure~\ref{fig:misclass-appx}. Furthermore, in Figure~\ref{fig:error_rdr_resnet}, we compare the instances within the RDR and the randomly chosen samples with the same label. In the first row of Figure~\ref{fig:error_rdr_resnet}, the dugong image is misclassified as the jellyfish. The instances in the RDR have faint objects compared to the general dugong cases. With such color schemes, instances usually belong to the jellyfish class. In this way, we can utilize the RDR framework misclassification reasoning. 

\section{Relaxed Decision Region with various settings}

This section supports the experimental results in the main paper. The Relaxed Decision Region best extracts the consistent but distinct features of the target instance. Even for datasets and models not visualized in the main paper, our framework could form the RDRs that find learned concepts well.

\subsection{Relaxed Decision Regions with Various Datasets}
\begin{figure}[ht] 
\centering
\centerline{\includegraphics[width=0.9\columnwidth]{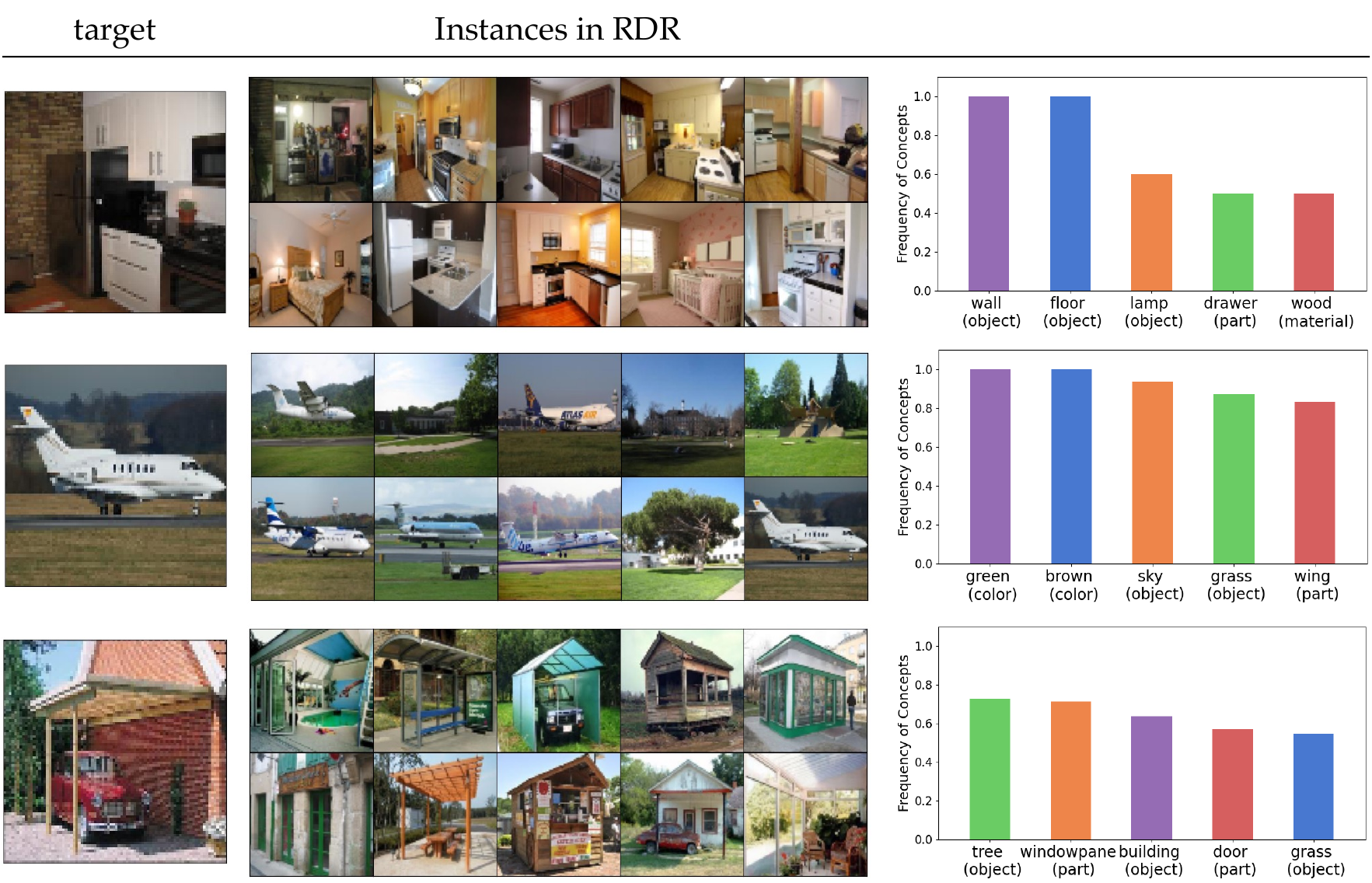}}
\caption{Relaxed Decision Region with Broden dataset, providing textual explanations of the concepts.} 
\label{fig:rdr_broden}
\end{figure}

\begin{figure}[ht] 
\centering
\centerline{\includegraphics[width=\columnwidth]{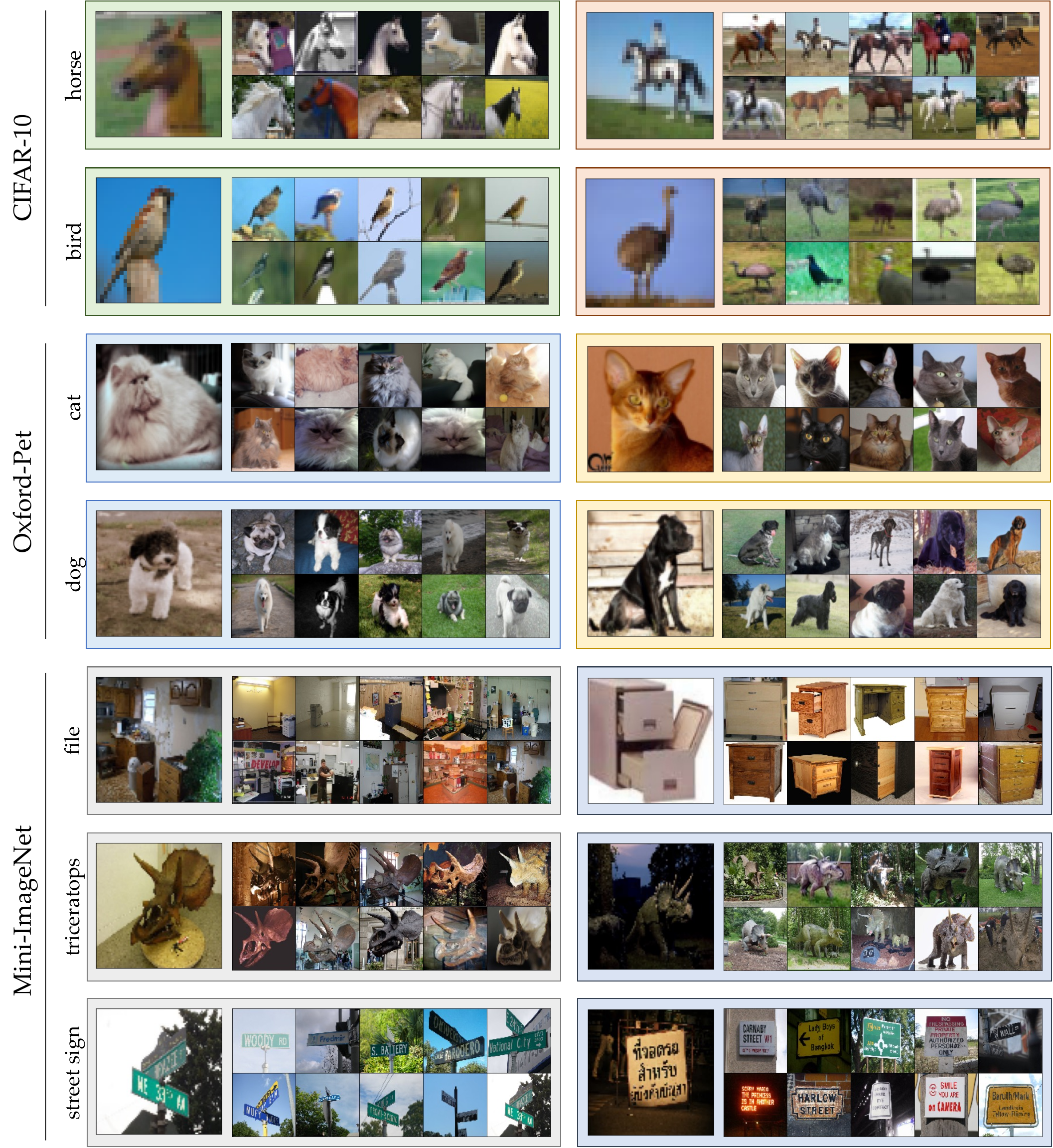}}
\caption{Relaxed Decision Region with CIFAR-10, Oxford-Pet and Mini-ImageNet dataset.} 
\label{fig:rdr_ex1}
\end{figure}

\begin{figure}[ht] 
\centering
\centerline{\includegraphics[width=\columnwidth]{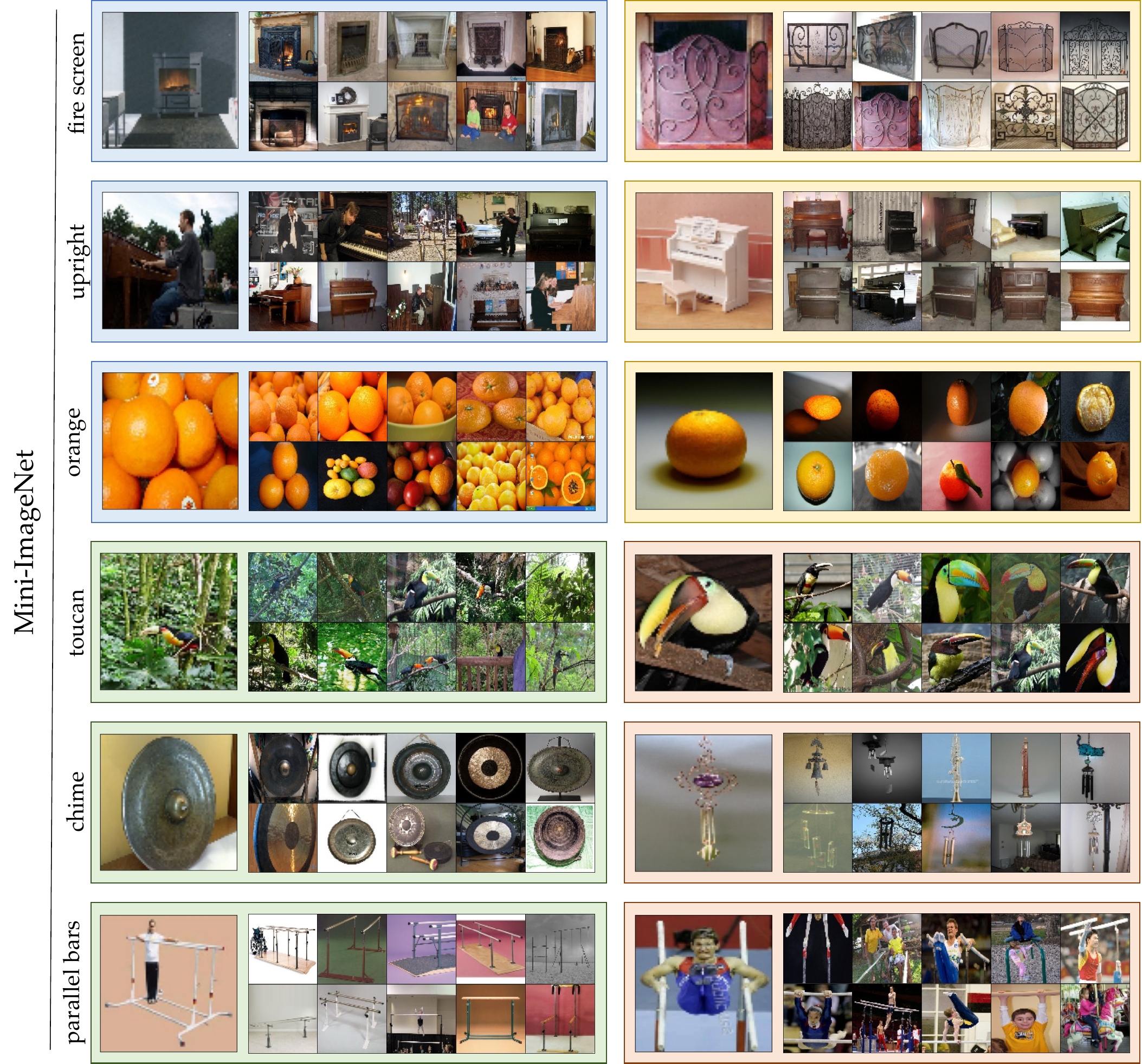}}
\caption{Relaxed Decision Region with CIFAR-10, Oxford-Pet and Mini-ImageNet dataset.} 
\label{fig:rdr_ex2}
\end{figure}

In order to evaluate the consistency of our algorithm, the Relaxed Decision Regions formed with various datasets are visualized in Figure~\ref{fig:rdr_ex1} and Figure~\ref{fig:rdr_ex2}. Furthermore, Figure~\ref{fig:rdr_broden} shows additional examples with the Broden dataset. The Broden dataset has pre-defined human-annotated concept labels, so we can check whether our algorithm captures the concept properly. We display the top 5 concept labels and their frequency within the RDR in Figure~\ref{fig:rdr_broden}.  The second-row instances in the RDR represent the concept `something horizontal, like an airplane on a green lawn', which are aligned with the Broden labels: green, brown, sky, grass and etc. 

\subsection{Relaxed Decision Regions with Various Models}

\begin{figure}[ht] 
\centering
\centerline{\includegraphics[width=\columnwidth]{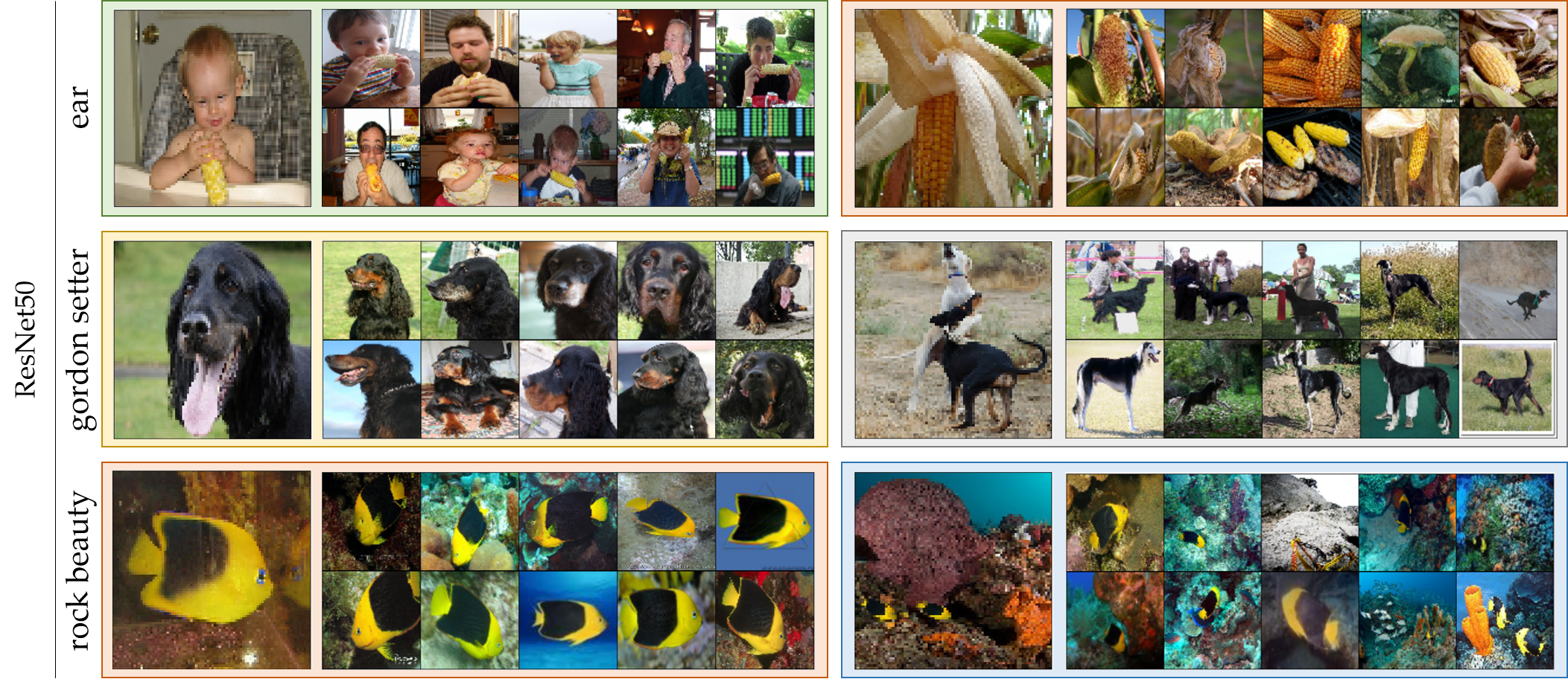}}
\caption{Relaxed Decision Region with ResNet50.} 
\label{fig:resnet}
\end{figure}

\begin{figure}[ht] 
\centering
\centerline{\includegraphics[width=\columnwidth]{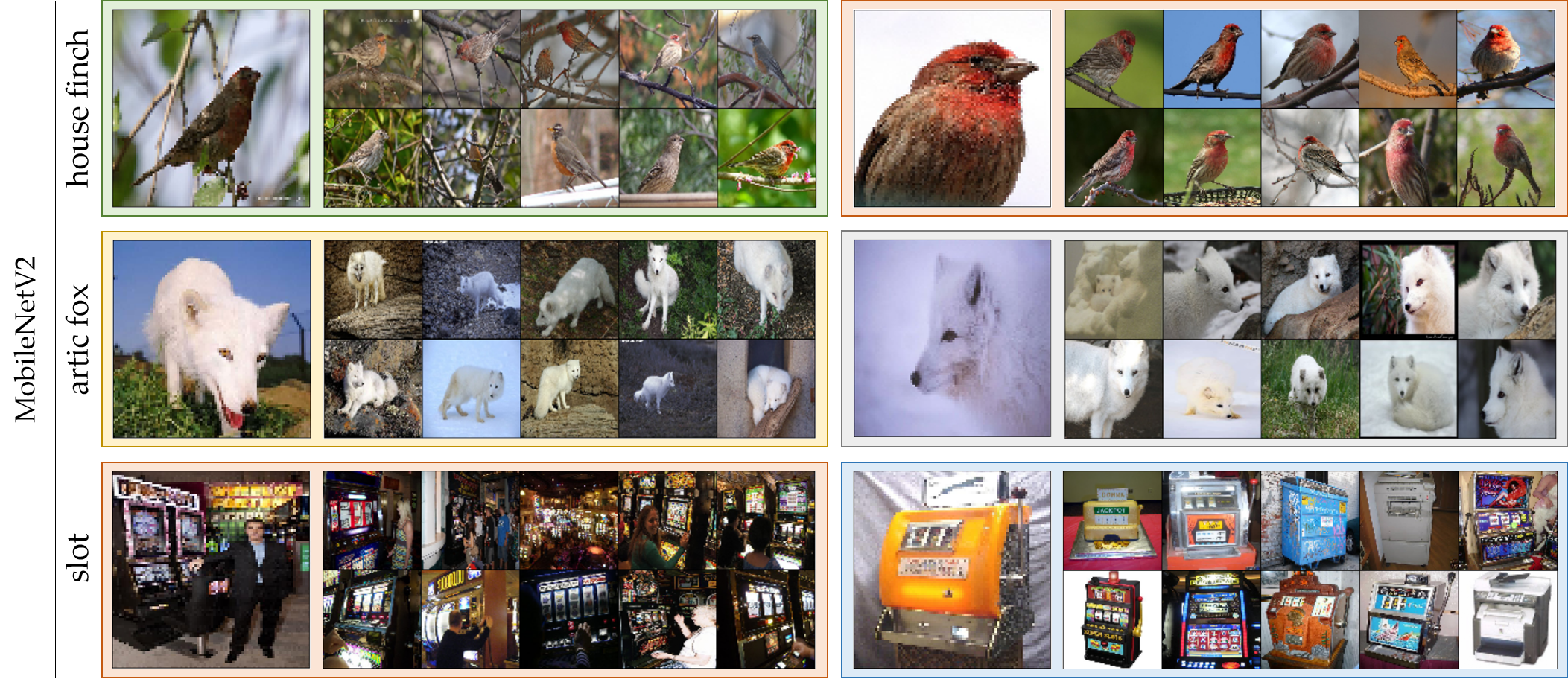}}
\caption{Relaxed Decision Region with MobileNetV2.} 
\label{fig:mobilenet}
\end{figure}

We construct the Relaxed Decision Regions with the Mini-ImageNet dataset on ResNet50 and MobileNetV2 as well as VGG19. In Figure~\ref{fig:resnet} and \ref{fig:mobilenet}, we identify that our algorithm has consistent performances on other deep neural network structures. 

\end{document}